\documentclass[letterpaper]{article} 
\usepackage{aaai2026}  
\usepackage{times}  
\usepackage{helvet}  
\usepackage{courier}  
\usepackage[hyphens]{url}  
\usepackage{graphicx} 
\usepackage{natbib}  
\usepackage{caption} 
\usepackage{algorithm}
\usepackage{algorithmic}

\usepackage{times}
\usepackage[utf8]{inputenc} 
\usepackage[T1]{fontenc}    
\usepackage{booktabs}       
\usepackage{amsfonts}       
\usepackage{nicefrac}       
\usepackage{microtype}      
\usepackage{xcolor}         

\usepackage{graphicx}
\usepackage[export]{adjustbox}
\usepackage{amsmath}
\usepackage{booktabs}
\usepackage{textcomp}
\usepackage{multirow} 
\usepackage{soul}
\usepackage{letltxmacro}
\usepackage{subcaption}
\usepackage{cleveref}

\usepackage{newfloat}
\usepackage{listings}
\DeclareCaptionStyle{ruled}{labelfont=normalfont,labelsep=colon,strut=off} 
\floatstyle{ruled}
\newfloat{listing}{tb}{lst}{}
\floatname{listing}{Listing}
\title{Object-Centric Latent Action Learning}
\author {
    Albina Klepach\textsuperscript{\rm 1},
    Alexander Nikulin\textsuperscript{\rm 1,\rm 2},
    Ilya Zisman\textsuperscript{\rm 1},
    Denis Tarasov\textsuperscript{\rm 1},
    Alexander Derevyagin\textsuperscript{\rm 1,\rm 3},
    Andrei Polubarov\textsuperscript{\rm 1}, 
    Nikita Lyubaykin\textsuperscript{\rm 1,\rm 4},
    Igor Kiselev\textsuperscript{\rm 5,},
    Vladislav Kurenkov\textsuperscript{\rm 1,\rm 4}
}
\affiliations {
    \textsuperscript{\rm 1}dunnolab.ai\\
    \textsuperscript{\rm 2}Moscow State University\\
    \textsuperscript{\rm 3}Higher School of Economics\\
    \textsuperscript{\rm 4}Innopolis University\\
    \textsuperscript{\rm 5}Accenture\\
    albina.klepach@gmail.com
}

\begin{document}

\maketitle

\begin{abstract}
Leveraging vast amounts of unlabeled internet video data for embodied AI is currently bottlenecked by the lack of action labels and the presence of action-correlated visual distractors. Although recent latent action policy optimization (LAPO) has shown promise in inferring proxy action labels from visual observations, its performance degrades significantly when distractors are present. To address this limitation, we propose a novel object-centric latent action learning framework that centers on objects rather than pixels. We leverage self-supervised object-centric pretraining to disentangle the movement of the agent and distracting background dynamics. This allows LAPO to focus on task-relevant interactions, resulting in more robust proxy-action labels, enabling better imitation learning and efficient adaptation of the agent with just a few action-labeled trajectories. We evaluated our method in eight visually complex tasks across the Distracting Control Suite (DCS) and Distracting MetaWorld (DMW). Our results show that object-centric pretraining mitigates the negative effects of distractors by \textbf{50\%}, as measured by downstream task performance: average return (DCS) and success rate (DMW).
\end{abstract}

\begin{links}
    \link{Code}{https://github.com/dunnolab/object-centric-lapo}
\end{links}

\section{Introduction}



In recent years, the scaling of model and data sizes has led to the creation of powerful and general foundation models \citep{bommasani2021opportunities} that have enabled many breakthroughs in understanding and generation of natural language \citep{achiam2023gpt, brown2020language} and images \citep{dehghani2023scaling, radford2021learning}. In contrast, the field of embodied AI has generally remained behind in terms of generalization and emergent abilities, being mostly limited by the lack of diverse high-quality data for pre-training \citep{guruprasad2024benchmarking, lin2024data}. The vast amount of video data on the Internet, covering a wide variety of human-related activities, can potentially fulfill the current data needs for training generalist agents \citep{mccarthy2024towards}. Unfortunately, such videos cannot be used directly due to lack of explicit action labels, which are essential for imitation learning and reinforcement learning algorithms.

\begin{figure}[t]
\centering
\includegraphics[width=\columnwidth]{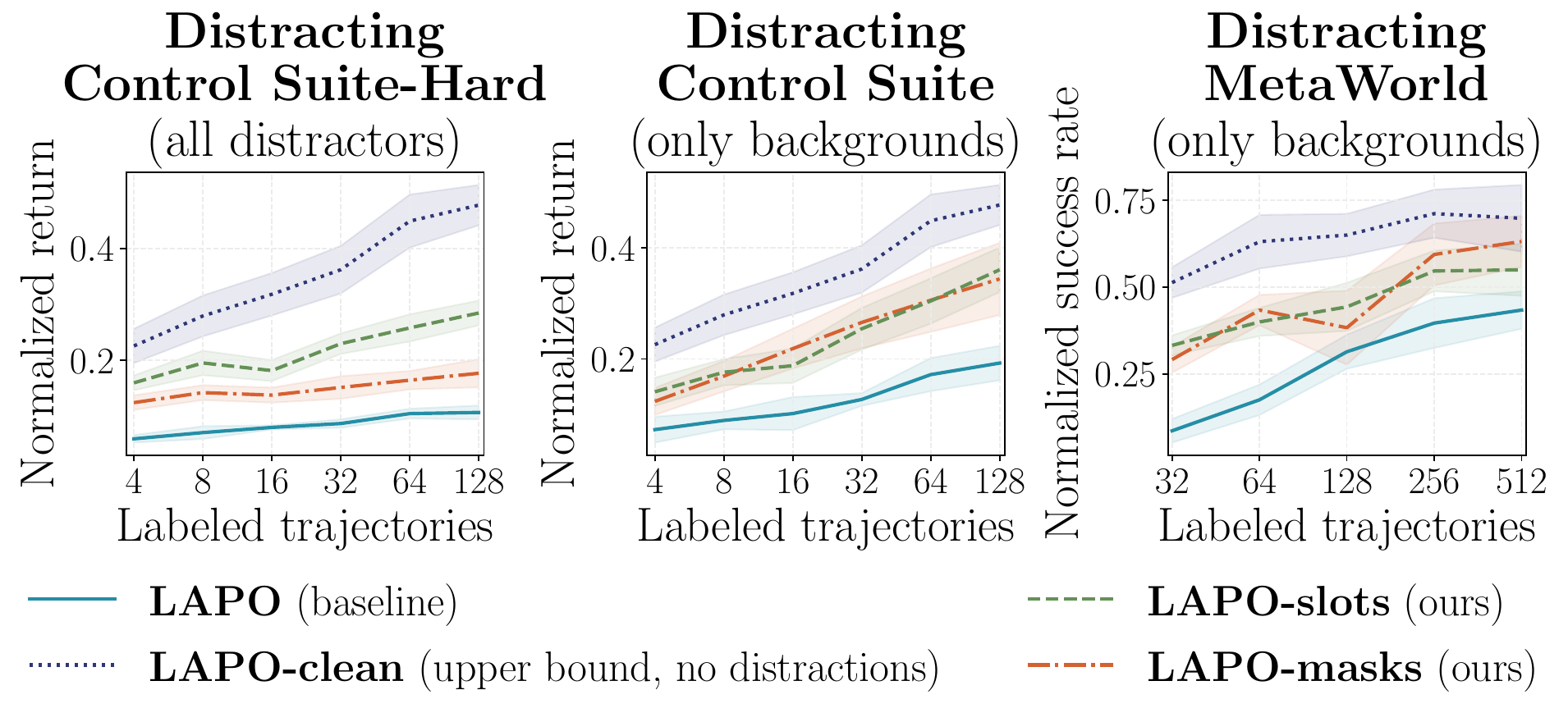}
\caption{\textbf{Main Results.} 
Our methods, \textbf{LAPO-slots} and \textbf{LAPO-masks}, leverage object-centric representations to significantly improve latent action learning under visual distractions. 
Compared to baseline (\textbf{LAPO} \citep{schmidt2024learningactactions}, trained on raw videos with distraction), our approaches reduce the performance gap toward the clean-data upper bound (\textbf{LAPO-clean}, trained on clean videos without distractions) by \textbf{50\%}. 
Distractions include camera movements, color variations, and dynamic background videos. 
Downstream performance is normalized to a behavior cloning agent trained on all ground-truth action labels. 
Results averaged over three random seeds. See \Cref{sec:exp} for details.
}
\label{fig:mean-performance}
\end{figure}

In order to compensate for the lack of action labels, approaches based on Latent Action Models (LAMs) \citep{schmidt2024learningactactions, ye2024latentactionpretrainingvideos, cui2024dynamo, bruce2024genie}, aim to infer latent actions from the sequence of visual observations. 
Such latent actions can later be used for behavioral cloning from large unlabeled datasets. 
However, a major limitation of current LAM approaches, is their susceptibility to action-correlated distractors, such as dynamic backgrounds, that falsely correlate with agent actions, and may lead models to overfit to non-causal patterns \citep{wang2024ad3, misra2024towards, mccarthy2024towards, nikulin2025latent}. 
Existing methods, such as Latent Action Pretraining (LAPA) \citep{ye2024latentactionpretrainingvideos, chen2024igor}, often assume curated, distractor-free datasets or rely on costly annotations, severely limiting their scalability and applicability in realistic settings.
\begin{figure*}[t]
\centering
\includegraphics[width=0.8\textwidth]{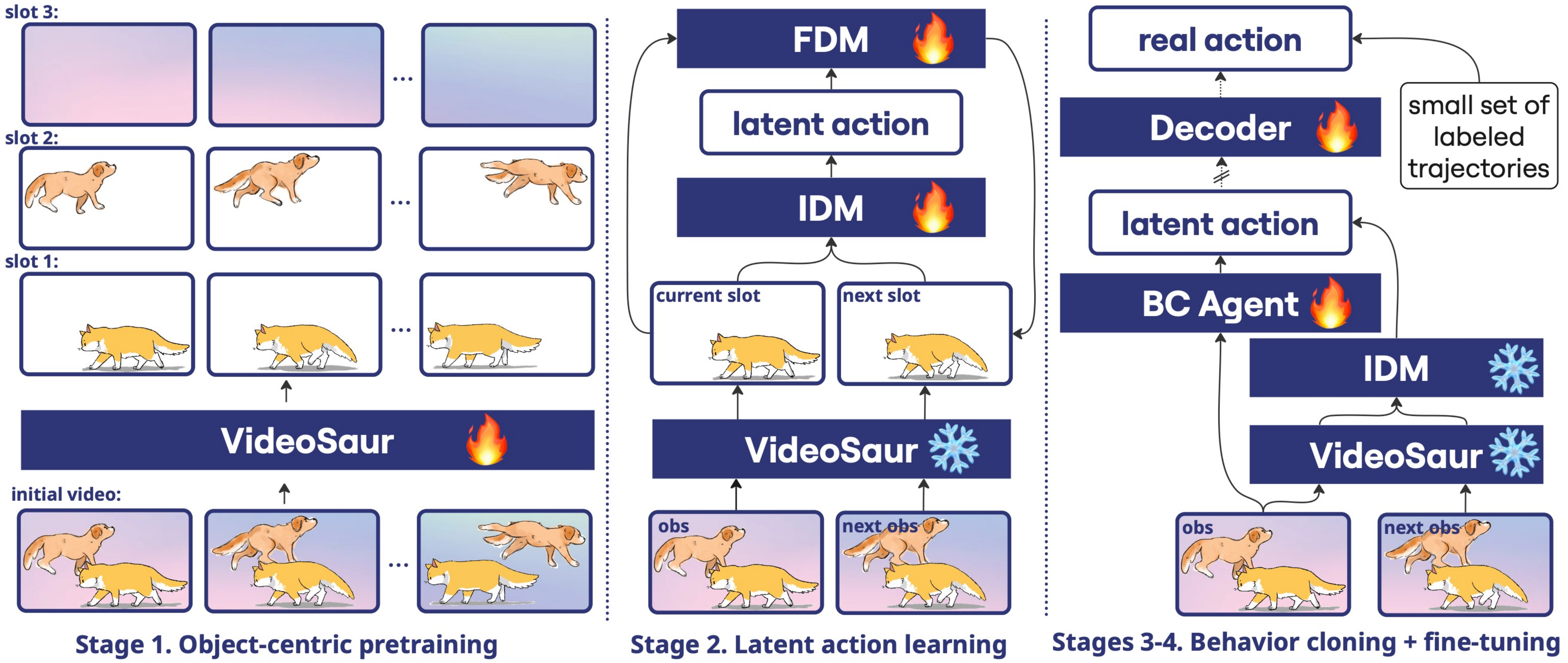}
\caption{General overview of our method pre-training pipeline. \textbf{Object-Centric Pretraining}: We decompose video sequences into interpretable object slots. A linear probe trained on slot representations automatically selects task-relevant slots by identifying those most predictive of actions. \textbf{Latent Action Learning}: We train a latent action model based on LAPO, learning inverse and forward dynamics in the slot space. \textbf{Behavior Cloning and Fine-tuning}: A behavior cloning agent is trained on the inferred latent actions. The resulting policy is then fine-tuned using a small number of trajectories with ground-truth actions ($\leq2.5$\% of total data), enabling strong downstream performance with minimal supervision.}
\label{fig:scheme}
\end{figure*}

To address this bottleneck, we propose object-centric latent action learning, a framework that leverages object-centric representations to isolate task-relevant entities from distracting visual noise. 
Object-centric models decompose scenes into discrete, interpretable object slots \citep{locatello2020object, dittadi2022generalization} using self-supervised mechanisms that group pixels into coherent entities based on feature similarity. These slots encode spatio-temporal properties \citep{zadaianchuk2023objectcentric} effectively disentangling complex visual inputs into semantically meaningful units.

Our core hypothesis is that object-centric representations provide the necessary structural priors to separate causal agent-object interactions from non-causal visual correlations. By operating on slot-based object features rather than raw pixels, our method enables LAMs to focus on the dynamics of relevant objects while filtering out distractions such as background motion.

We evaluated our approach in continuous control environments with dynamic distractors, such as Distracting Control Suite (DCS) \citep{stone2021distractingcontrolsuite} and previously unexplored Distracting Metaworld (DMW) \citep{yu2019metaworld}, showing that self-supervised object-centric pretraining, built on VideoSAUR \citep{zadaianchuk2023objectcentric}, significantly improves robustness compared to standard LAM baselines. Our results suggest that object decomposition serves as an effective inductive bias for latent action learning, especially under realistic, noisy conditions. In addition, we propose a simple way to automatically choose slots relevant to control and analyze its effectiveness. We summarize our main results on \Cref{fig:mean-performance}.

\section{Preliminaries}

In our experiments, we use environments with partial observability (POMDP) \citep{aastrom1965optimal, kaelbling1998planning}. A POMDP is defined by the tuple $(\mathcal{S}, \mathcal{A}, T, R, \Omega, \gamma)$, where: $\mathcal{S}$ is the state space, $\mathcal{A}$ is the action space, $T(s_{t+1} \mid s_t, a_t)$ defines the transition dynamics, $R(s_t, a_t, s_{t+1})$ is the reward function, $\Omega(o_t \mid s_t)$ is the observation model, $\gamma \in [0,1)$ is the discount factor. At each time step $t$, the agent receives an observation, $o_t \sim \Omega(\cdot \mid s_t)$, and selects an action $a_t \sim \pi(\cdot \mid o_t)$ to maximize its expected discounted return: $\mathbb{E}\left[\sum_{k=t}^\infty \gamma^{k-t} r_k \right]$ The agent transitions to the next state $s_{t+1} \sim T(\cdot \mid s_t, a_t)$ and receives a reward $r_t = R(s_t, a_t, s_{t+1})$.

We consider the setting in which we first learn a policy $\pi$ from offline actionless demonstration data, then fine-tune it on real actions, introduced by \cite{schmidt2024learningactactions}. Suppose we are given a dataset $\mathcal{D}$ of expert trajectories in the form of sequences of observations and actions:  
$$
\tau_{oa} = \{(o_0, a_0, o_1), ..., (o_{|\tau|-1}, a_{|\tau|-1}, o_{|\tau|})\}
$$

When such labeled data is available, Behavior Cloning (BC) \citep{pomerleau1988alvinn} can be used to train a policy $\pi_{\text{BC}}(a_t \mid o_t)$ through supervised learning. However,  when using real-world video data the action labels are missing, and we only observe sequences of the form:
$$
\tau_{o} = \{o_0, o_1, ..., o_{|\tau|}\}
$$

\paragraph{Latent Action Modeling.} \citet{edwards2019imitating, schmidt2024learningactactions} proposed to convert classic BC into a pseudo-BC task by inferring latent actions $\mathbf{z}_t$ that explain each observed transition $(o_t, o_{t+1})$. These inferred actions are then used to train a latent action policy $\tilde{\pi}(z_t \mid o_t)$, bypassing the need for ground-truth action labels. To obtain latent actions, LAM trains an inverse-dynamics model (IDM) $f_\mathrm{IDM}(\boldsymbol{z}_t|\boldsymbol{o}_t, \boldsymbol{o}_{t+1})$, which infers latent actions from state transitions. To ensure that the IDM learns meaningful representations, it is trained jointly with a forward-dynamics model (FDM): $f_\mathrm{FDM}(\boldsymbol{o}_{t+1}|\boldsymbol{o}_{t}, \boldsymbol{z}_{t})$, which predicts the next observation from the current observation and the inferred action. The models are optimized to perform next observation prediction:
$$
\mathcal{L}_\mathrm{MSE} = \mathbb{E}_{t} \left[ \left\| f_\mathrm{FDM}(f_\mathrm{IDM}(\boldsymbol{o}_t, \boldsymbol{o}_{t+1}), \boldsymbol{o}_t) - \boldsymbol{o}_{t+1} \right\|^2 \right]
$$

Without distractors LAM is able to recover ground-truth actions \citep{schmidt2024learningactactions, bruce2024genie}, however, with distractions it does not work \citep{nikulin2025latent}.

\paragraph{Challenge of distractors.} 
Real-world videos inherently contain action-correlated distractors: environmental dynamics (e.g. moving backgrounds, camera jitter) that spuriously correlate with agent actions. However, existing learning from observation methods lack mechanisms to disentangle distractors \citep{efroni2022provablerlexogenousdistractors, misra2024principledrepresentationlearningvideos}, leading to overfitting in noisy settings. 
Recently, \citet{nikulin2025latent} empirically demonstrated that LAPO-based methods suffer significant performance degradation when trained on data that contain distractors and suggested reusing available action labels to provide supervision during LAM pre-training. However, it is not always possible to provide such supervision, since action labels do not exist in principle in some domains (e.g. YouTube videos). On the other hand, object-centric decomposition can be applied in a meaningful way to any real-world video data, greatly expanding the applicability of latent action learning. 

\paragraph{Object-Centric Decomposition.} Object-centric learning (OCL) aims to decompose complex visual scenes into structured, interpretable representations by isolating individual entities over time. These entities are typically encoded as a set of $K$ slot vectors $\mathcal{S}_t = \{\boldsymbol{s}_t^{(1)}, \boldsymbol{s}_t^{(2)}, ..., \boldsymbol{s}_t^{(K)}\},$ representing a distinct entity or region within the scene and capture object-specific properties such as position, shape, color, and motion. 
A key strength of OCL models lies in their ability to encode the causal structure of the environment. By isolating coherent, temporally consistent entities, they provide a natural defense against distractors in real-world video data. For example, VideoSAUR \citep{zadaianchuk2023objectcentric}, a self-supervised video decomposition model, has been shown to effectively ignore spurious correlations introduced by moving backgrounds or viewpoint changes, focusing instead on task-relevant objects.

Although object slots provide disentangled and semantically meaningful representations, identifying which slots correspond to task-relevant entities remains challenging and necessitates additional mechanisms to identify and track relevant entities without manual supervision. 
Also, most OCL frameworks, require the number of slots $K$ to be specified a priori, which limits flexibility in modeling scenes with varying numbers of objects and can lead to either under-representation or fragmentation when $K$ is mismatched with scene complexity. 




\begin{figure}[t]
\centering
\includegraphics[width=0.8\columnwidth]{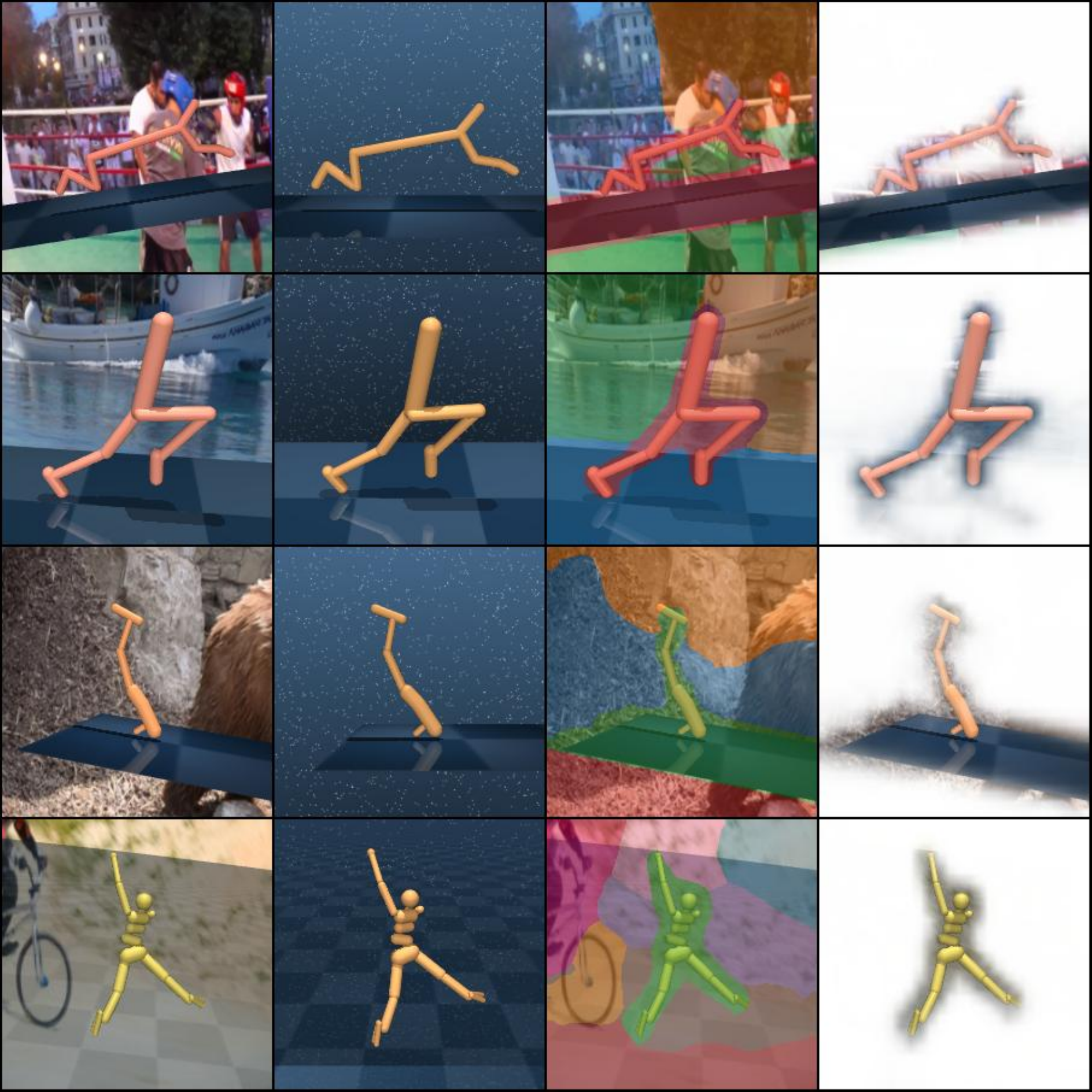}
\caption{Visuals from Distraction Control Suite. From top to bottom: cheetah-run, walker-run, hopper-hop, humanoid-walk. From left to right: the distracted observation (background video, color, and camera position variations), the non-distracted observation, the mixture of slot decoder masks obtained after object-centric pretraining, and the main object slot decoder mask selected after object-centric pretraining.}
\label{fig:slot_examples}
\end{figure}

\section{Method} 
\label{section:oc-lapo}


\paragraph{Object-Centric Representation Learning.} We employ the VideoSAUR model \citep{zadaianchuk2023objectcentric} to decompose input video frames into spatio-temporal object slots.  We initially experimented with the STEVE model \citep{singh2022simpleunsupervisedobjectcentriclearning}, a widely used and promising approach, but found that it failed to consistently isolate entities such as the hopper in our tasks (see Supplementary Material).
VideoSaur's self-supervised architecture isolates individual entities within a scene, providing structured representations that are less susceptible to background noise and incidental motion. At the end of this step, we obtain an encoder, that directly maps a trajectory of observations $\tau_{o} = ({o}_1, ..., {o}_T)$ to a corresponding trajectory of slots $\tau_{s} = ({s}_1, ..., {s}_T)$. VideoSAUR encodes each observation ${o}_t$ in a fixed number $K$ of slot vectors ${s}_t^{(k)} \in \mathbb{R}^d$, where the parameter $K$ is set prior to training.
Due to its transformer-based decoder, VideoSAUR allows each slot ${s}_t^{(k)}$ to be projected back into the original observation space using attention maps as alpha masks. This enables us to visualize and interpret the spatial support of each slot in the image. We denote these by ${m}_t^{(k)}$ and refer to them as object masks throughout the paper (see \Cref{fig:slot_examples} and Supplementary Material for visualizations).
To improve consistency between training runs and reduce slot permutation issues, we employ fixed initialization of slots (see details in Supplementary Material), which ensures that the same slot index corresponds to a similar semantic object across different episodes.

\begin{table*}[t]
\begin{center}
\begin{tabular}{c|l|c|c|rlrl}
\toprule
    \bf Environment \& 
    & \multicolumn{1}{|c|}{\bf Task \&}
    & \bf LAPO
    & \bf{LAPO-clean}
    & \multicolumn{2}{c}{\textbf{LAPO-masks}} 
    & \multicolumn{2}{c}{\textbf{LAPO-slots}} \\
    \bf Metric
    & \bf mean over tasks
    & baseline 
    & upper-bound 
    & \multicolumn{1}{|c}{ours} 
    & \multicolumn{1}{c}{+\textit{gain}\%} 
    & \multicolumn{1}{c}{ours} 
    & \multicolumn{1}{c}{+\textit{gain}\%} \\
\midrule

\multirow{5}{*}{\shortstack[c]{\textbf{DCS-Hard} \\ \textit{(Normalized Return)}}}
    & cheetah-run & 0.24 {\scriptsize ± 0.02} & 0.76 {\scriptsize ± 0.04} & 0.41 {\scriptsize ± 0.03} & +32\% & 0.55 {\scriptsize ± 0.04} & +58\% \\
    & hopper-hop  & 0.03 {\scriptsize ± 0.01} & 0.27 {\scriptsize ± 0.03} & 0.08 {\scriptsize ± 0.01} & +20\% & 0.15 {\scriptsize ± 0.02} & +50\% \\
    & walker-run  & 0.04 {\scriptsize ± 0.01} & 0.32 {\scriptsize ± 0.07} & 0.06 {\scriptsize ± 0.01} & +6\%  & 0.12 {\scriptsize ± 0.01} & +27\% \\
    & humanoid-walk & 0.02 {\scriptsize ± 0.01} & 0.06 {\scriptsize ± 0.01} & 0.04 {\scriptsize ± 0.02} & +47\% & 0.06 {\scriptsize ± 0.01} & +105\% \\
\cmidrule(l){2-8}
    & \textbf{mean} & 0.08 {\scriptsize ± 0.01} & 0.35 {\scriptsize ± 0.04} & 0.15 {\scriptsize ± 0.02} & \textbf{+26\%} & 0.22 {\scriptsize ± 0.02} & \textbf{+52\%} \\
\midrule

\multirow{5}{*}{\shortstack[c]{\textbf{DCS} \\ \\ \textit{(Normalized Return)}}}
    & cheetah-run & 0.34 {\scriptsize ± 0.06} & 0.76 {\scriptsize ± 0.04} & 0.54 {\scriptsize ± 0.09} & +48\% & 0.50 {\scriptsize ± 0.07} & +38\% \\
    & hopper-hop  & 0.05 {\scriptsize ± 0.01} & 0.27 {\scriptsize ± 0.03} & 0.15 {\scriptsize ± 0.01} & +46\% & 0.19 {\scriptsize ± 0.02} & +64\% \\
    & walker-run  & 0.10 {\scriptsize ± 0.01} & 0.32 {\scriptsize ± 0.07} & 0.21 {\scriptsize ± 0.05} & +52\% & 0.18 {\scriptsize ± 0.03} & +36\% \\
    & humanoid-walk & 0.02 {\scriptsize ± 0.01} & 0.06 {\scriptsize ± 0.01} & 0.05 {\scriptsize ± 0.02} & +67\% & 0.09 {\scriptsize ± 0.02} & +174\% \\
\cmidrule(l){2-8}
    & \textbf{mean} & 0.13 {\scriptsize ± 0.02} & 0.35 {\scriptsize ± 0.04} & 0.24 {\scriptsize ± 0.04} & \textbf{+50\%} & 0.24 {\scriptsize ± 0.03} & \textbf{+50\%} \\
\midrule

\multirow{5}{*}{\shortstack[c]{\textbf{DMW} \\ \textit{(Normalized Success Rate)}}}
    & hammer & 0.75 {\scriptsize ± 0.07} & 0.98 {\scriptsize ± 0.01} & 0.96 {\scriptsize ± 0.01} & +91\% & 0.99 {\scriptsize ± 0.02} & +102\% \\
    & bin-picking & 0.18 {\scriptsize ± 0.08} & 0.74 {\scriptsize ± 0.10} & 0.49 {\scriptsize ± 0.10} & +56\% & 0.33 {\scriptsize ± 0.08} & +27\% \\
    & basketball & 0.17 {\scriptsize ± 0.03} & 0.51 {\scriptsize ± 0.07} & 0.34 {\scriptsize ± 0.09} & +50\% & 0.37 {\scriptsize ± 0.09} & +58\% \\
    & soccer & 0.14 {\scriptsize ± 0.06} & 0.36 {\scriptsize ± 0.08} & 0.21 {\scriptsize ± 0.08} & +34\% & 0.23 {\scriptsize ± 0.06} & +41\% \\
\cmidrule(l){2-8}
    & \textbf{mean} & 0.31 {\scriptsize ± 0.06} & 0.65 {\scriptsize ± 0.06} & 0.50 {\scriptsize ± 0.07} & \textbf{+55\%} & 0.48 {\scriptsize ± 0.06} & \textbf{+50\%} \\
\bottomrule
\end{tabular}
\end{center}

\caption{Object-centric pretraining substantially improves robustness to visual distractors, where standard imitation learning from observation methods like LAPO struggle. In environments with visual distractors: dynamic background videos (DCS and DMW) and additional camera motion and color variation (DCS-Hard), LAPO's performance is significantly degraded compared to the clean-data upper bound. We report normalized evaluation returns (DCS) and success rates (DMW) for a behavior cloning (BC) agent trained on latent actions, where a score of 1.0 corresponds to a BC agent trained on the full set of ground-truth actions. Each value averages performance across varying amounts of fine-tuning data and three random seeds. The relative performance gain is defined as:
$
\frac{\text{LAPO-slots/masks}\,\, -\,\,  \text{LAPO}}{\text{LAPO-clean}\,\, -\,\, \text{LAPO}} \times 100\%
$.
This measures the fraction of the recoverable performance gap between the distracted baseline (LAPO) and the clean-data performance (LAPO-clean) that our method recovers. LAPO-slots and LAPO-masks achieve large improvements over LAPO, closing up to $\sim 50\%$ of the gap, demonstrating that object-centric representations effectively mitigate the negative effects of visual distractions.}
\label{table:eval-returns}
\end{table*}

\paragraph{Slot Selection via Linear Action Probe.}
In our setting, control-relevant entities typically include the main agent (e.g., the cheetah in DCS tasks or the robotic arm and gripper in DMW) and task-critical objects (e.g., hammer, ball, or cubes). Depending on the environment and the number of slots $K$, these entities may be encoded across one or more slots. To automatically identify which slots capture action-relevant information, we adopt a probing approach inspired by \citet{alain2016understanding}, measuring how well individual slot representations predict ground-truth actions.

We first apply Principal Component Analysis (PCA) to slot encodings computed from a small set of labeled trajectories, reducing dimensionality and noise. Then, a linear regressor is trained on resulting representations to predict the true action. To ensure robustness, we evaluate performance using 5-fold cross-validation and report the average test Mean Squared Error (MSE), which we refer to as the \textbf{Linear Action Probe score}. Lower MSE indicates stronger action predictivity and thus higher relevance to control.

We select the most predictive slots as the final set of relevant slots $\mathcal{S}^\star = \{\boldsymbol{s}^{(k)} \mid k \in \mathcal{K}^\star\}$, where $\mathcal{K}^\star$ denotes the selected slot indices. This selection is performed once per dataset after OC pretraining, leveraging the fixed slot initialization, ensuring stable and consistent slot interpretation across episodes.

\paragraph{Latent Action Modeling.} Utilizing the selected object-centric representations, we train a latent action model inspired by LAPO \citep{schmidt2024learningactactions}. 

\textbf{LAPO-slots} operates purely in the latent space. Its inverse-dynamics model $\boldsymbol{z}_t \sim f^{s}_\mathrm{IDM}(\cdot|\boldsymbol{s}_t, \boldsymbol{s}_{t+1})$ and forward-dynamics model $\boldsymbol{\hat{s}}_{t+q} \sim f^{s}_\mathrm{FDM}(\cdot|\boldsymbol{s}_{t}, \boldsymbol{z}_{t})$ are trained to reconstruct the next slot embeddings, minimizing 
 ${\vert \vert \boldsymbol{\hat{s}}_{t+1}-\boldsymbol{s}_{t+1}\vert \vert}^2$. 
 
 \textbf{LAPO-masks} operates in pixel space. It first creates a filtered image by applying the object masks from the selected slots to the input frame. The dynamics models are then trained to reconstruct this filtered image at the next timestep. Visualizations of the filtered images can be found in \Cref{fig:slot_examples,fig:ablations-B} and the Supplementary Material.

\begin{figure*}[t]
\centering

\centerline{\includegraphics[width=0.8\textwidth]{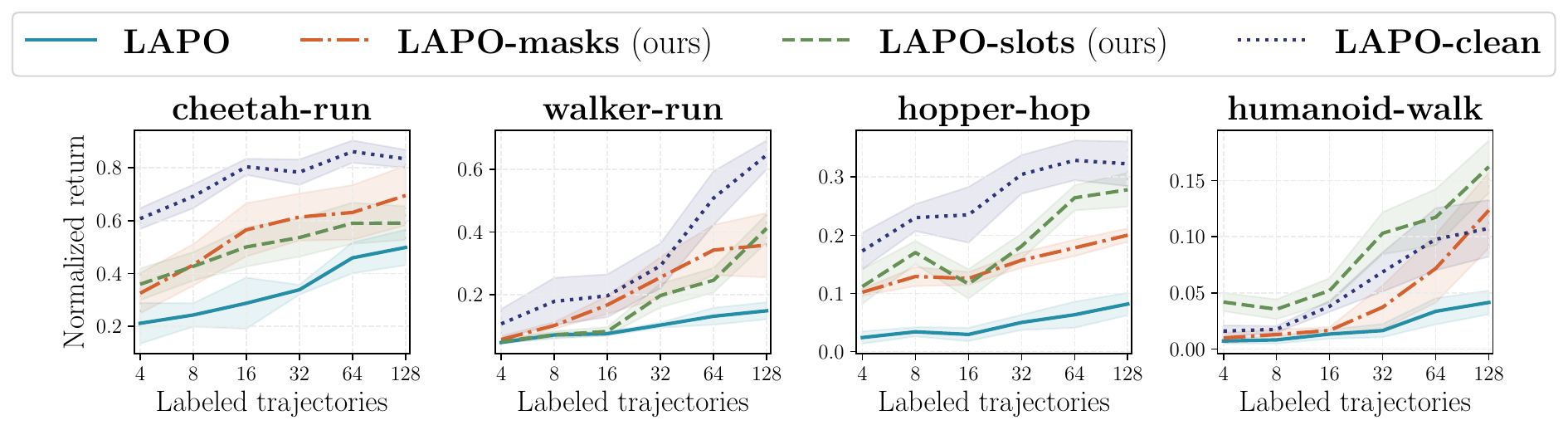}}
\centerline{\includegraphics[width=0.75\textwidth]{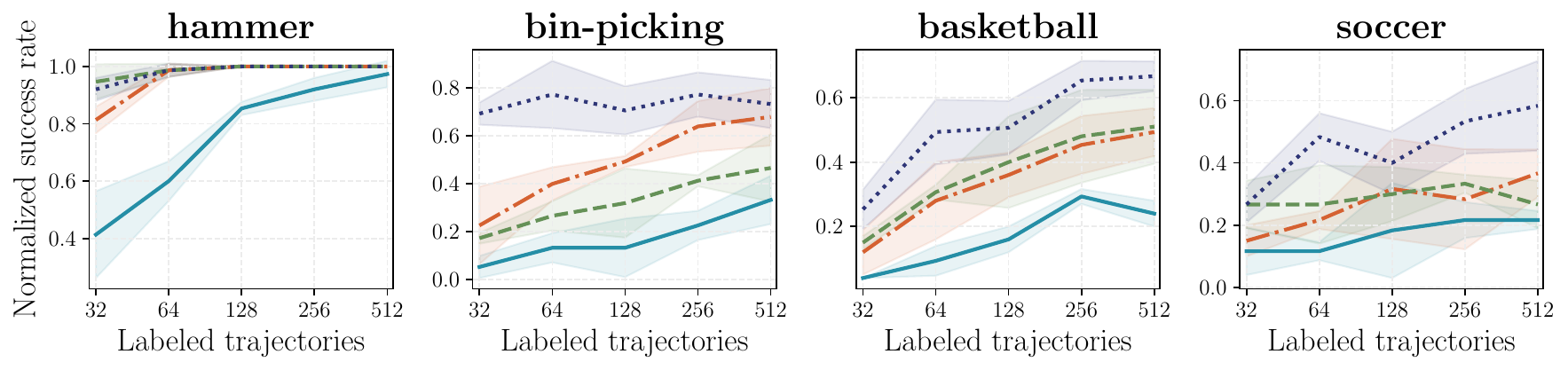}}

\caption{
Normalized evaluation returns and success rates of the BC agent trained on latent actions for varying numbers of fine-tuning labeled trajectories. 
TL;DR: Object-centric learning improves evaluation returns in DCS tasks and success rate in goal-based DMW task for all tasks.
The plots are arranged from left to right in order of increasing task complexity. The values are averaged across three random seeds.
The BC agent trained with access to the full dataset of ground-truth actions would return a score of 1 for each task. 
}
\label{fig:eval-returns}
\end{figure*}

\paragraph{Behavior Cloning and Finetuning.} The inferred latent actions serve as proxies for actual action labels. We train a behavior cloning (BC) agent to predict these latent actions, using the same dataset as for latent action learning. To evaluate the pre-training effectiveness, as a final stage, we fine-tune the resulting agent on a limited set of trajectories with ground-truth action labels (no more than 2.5\% of total data), in line with \citep{schmidt2024learningactactions, ye2024latentactionpretrainingvideos, nikulin2025latent}. The finetuned BC agent scores for different numbers of labeled trajectories are presented on \Cref{fig:eval-returns}. 

\section{Experiments} \label{sec:exp}

Our experiments are designed to address the following core questions:
Can object-centric representations disentangle task-relevant motion from distracting visual elements? 
How much supervision is needed to train effective latent action models under high visual distraction?
Can we automate slot selection to reduce the dependency on domain expertise or manual annotation?

\subsection{Datasets}
We evaluate on two benchmark environments: Distracting Control Suite (DCS) and Distracting Meta-World (DMW). Details on data collection are in Supplementary Material.

\paragraph{Distracting Control Suite (DCS).}
DCS~\citep{stone2021distractingcontrolsuite} extends DeepMind Control with three visual distractions: (1) dynamic backgrounds (DAVIS 2017 videos), (2) color variations (hue/saturation shifts), and (3) camera perturbations. We focus on dynamic backgrounds (\textbf{DCS}) and report full results on all distractions (\textbf{DCS-hard}) in \Cref{fig:mean-performance,table:eval-returns} and Supplementary Material. Tasks: cheetah-run, walker-run, hopper-hop, humanoid-walk (increasing complexity). Experts were trained on privileged state; datasets consist of observation-action pairs. Behavior cloning, trained on privileged true actions, is able to achieve expert performance.

\paragraph{Distracting Meta-World (DMW).}
DMW extends Meta-World~\citep{yu2019metaworld} with DCS-style dynamic backgrounds. Tasks (hammer, bin-picking, basketball, soccer) involve multi-object interactions and compositional reasoning. Experts use Meta-World’s oracle policies.

\begin{figure*}[t]
\centering
\def\myheight{4.2cm}
\def\myheightt{4.6cm}

\begin{subfigure}[t]{0.32\textwidth}
  \centering
  \includegraphics[height=\myheight]{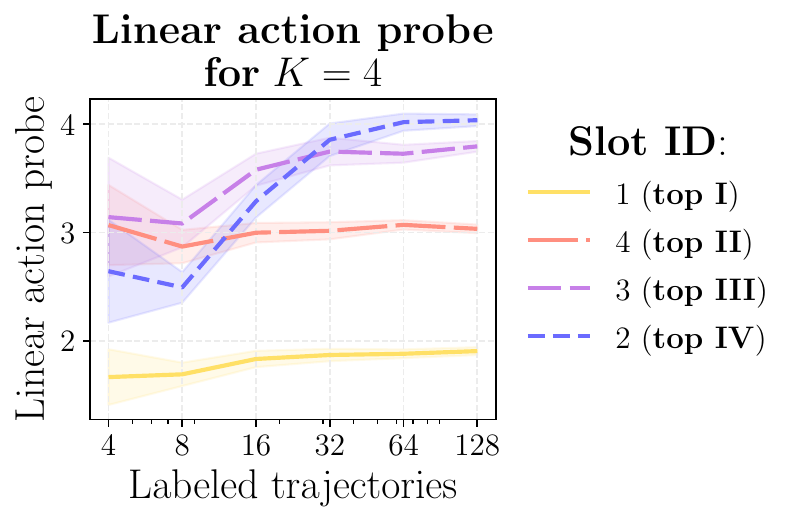}
  \caption{}
  \label{fig:ablations-A}
\end{subfigure}
\hfill
\begin{subfigure}[t]{0.32\textwidth}
  \centering
  \includegraphics[height=\myheight]{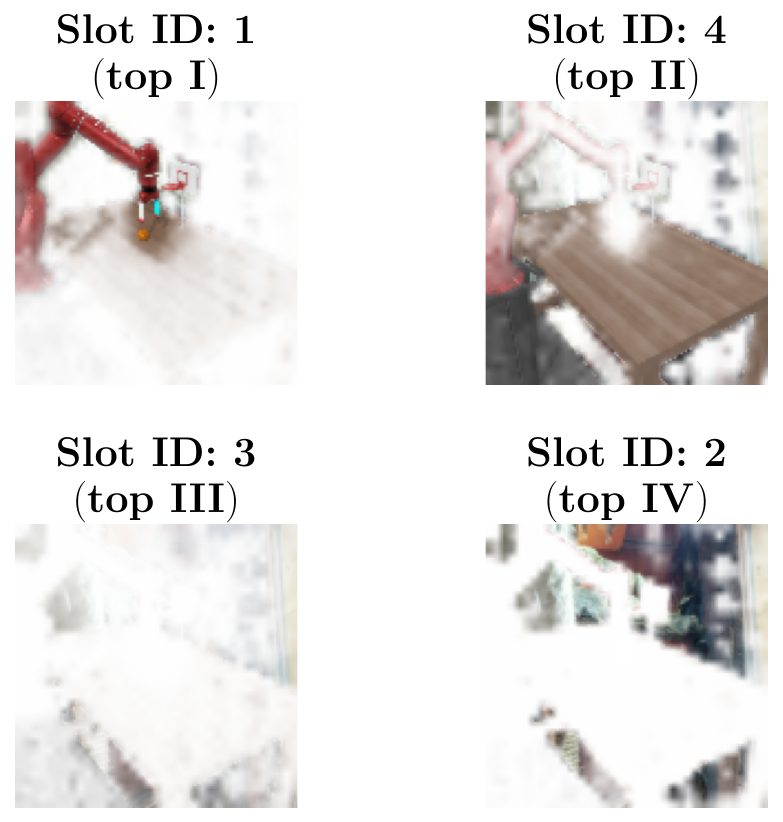}
  \caption{}
  \label{fig:ablations-B}
\end{subfigure}
\hfill
\begin{subfigure}[t]{0.32\textwidth}
  \centering
  \includegraphics[height=\myheight]{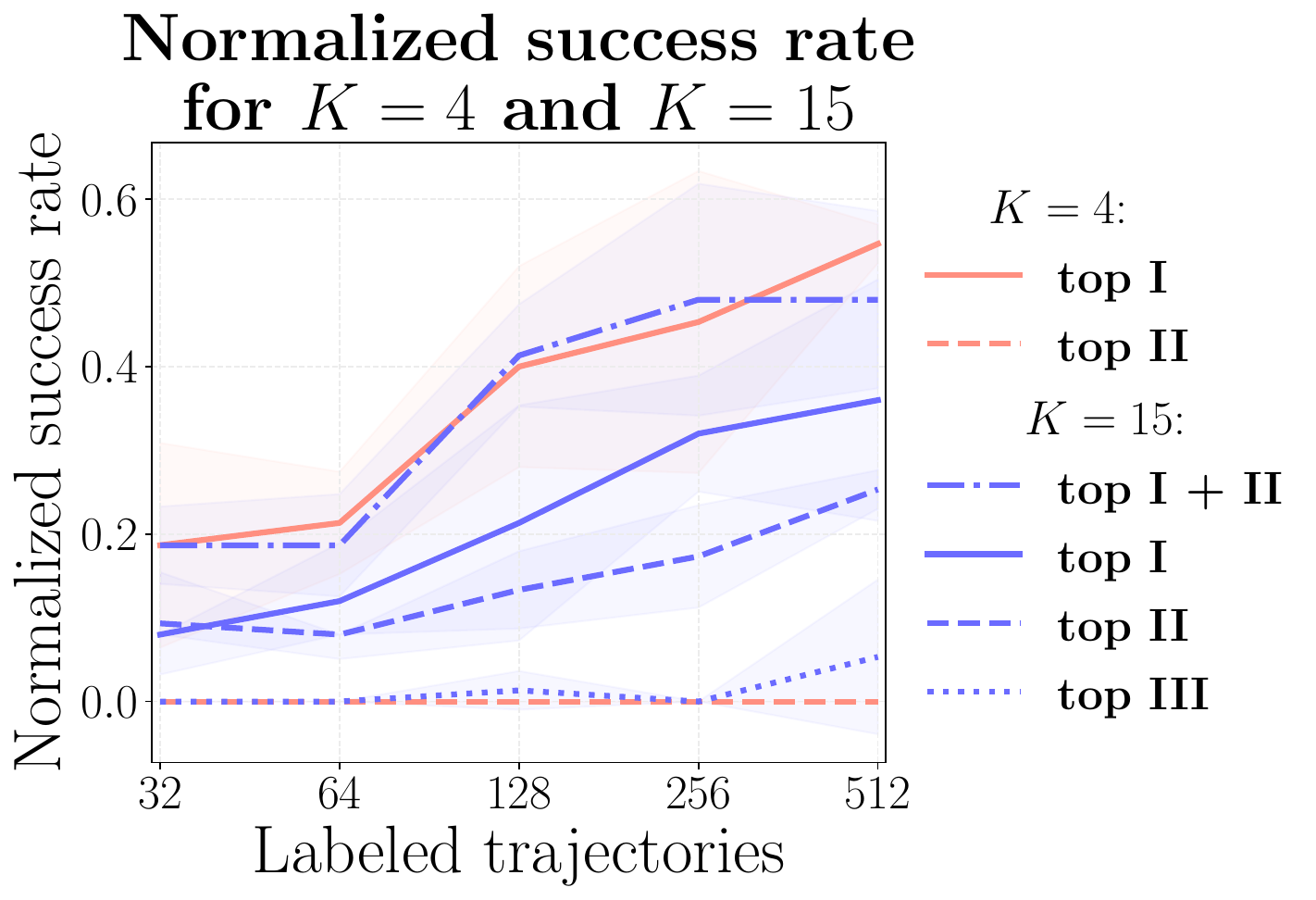}
  \caption{}
  \label{fig:ablations-C}
\end{subfigure}

\vspace{1.5ex}

\begin{subfigure}[t]{0.49\textwidth}
  \centering
  \includegraphics[height=\myheightt]{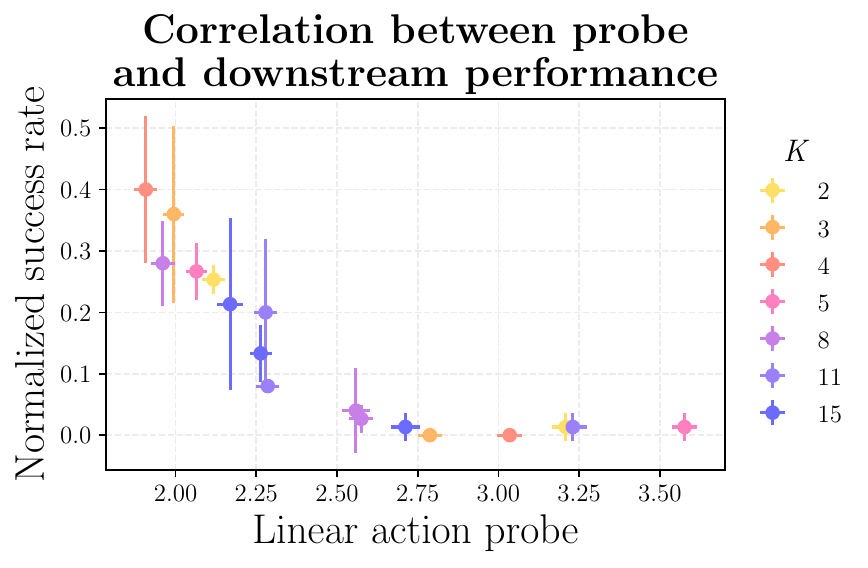}
  \caption{}
  \label{fig:ablations-D}
\end{subfigure}
\hfill
\begin{subfigure}[t]{0.49\textwidth}
  \centering
  \includegraphics[height=\myheightt]{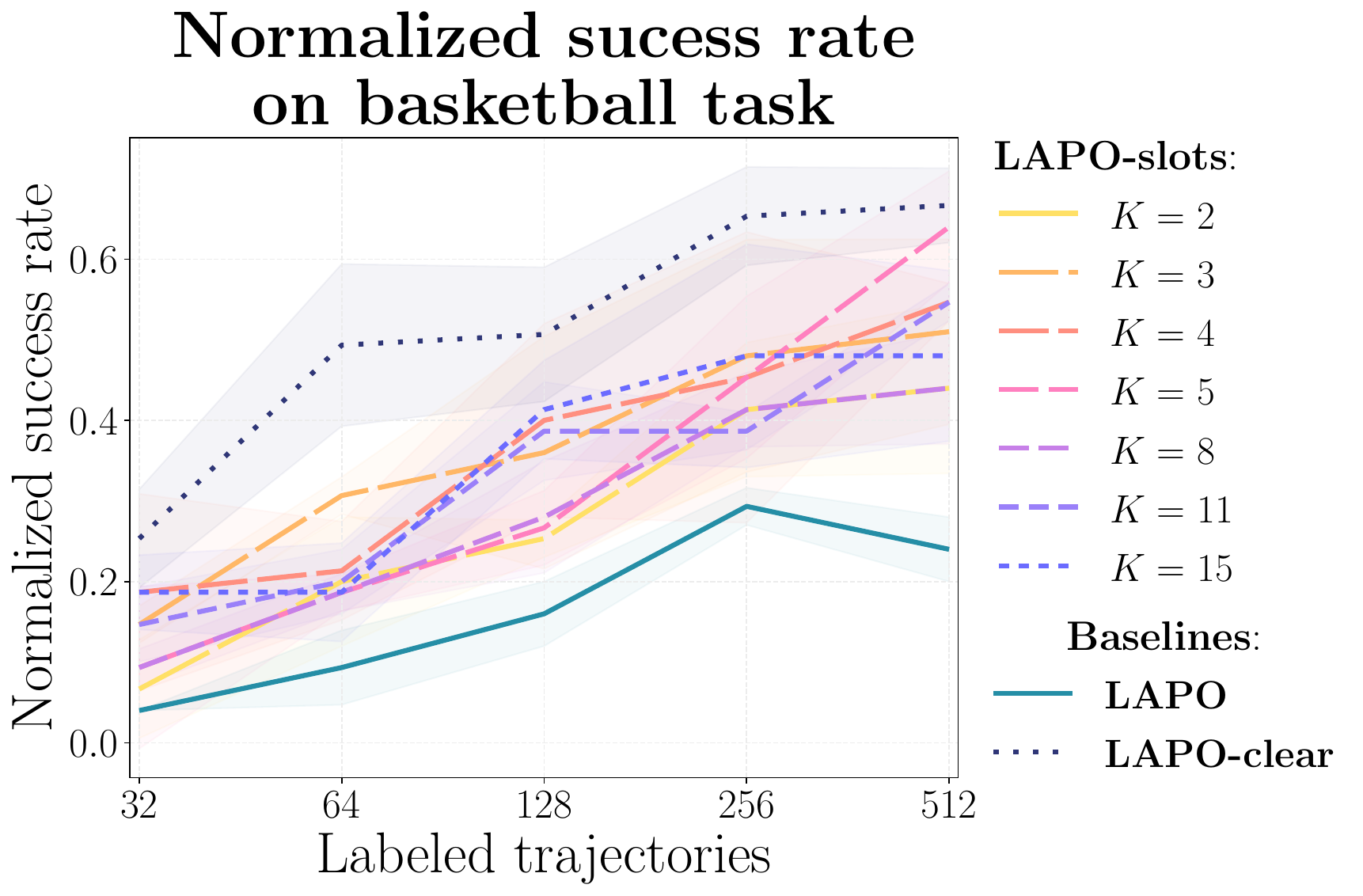}
  \caption{}
  \label{fig:ablations-E}
\end{subfigure}

\caption{
Slot selection study on basketball task.
(a,b) Linear action probes under varying labeled trajectories budget and corresponding examples of slot masks for basketball task for $K=4$.
(c) BC performance on different slots for $K=4$ and $K=15$ on basketball task.
(d) Linear action probe scores vs.\ normalized BC success rates (on 128 trajectories). TL;DR: Probe-based selection correlates with downstream performance. (e) BC performance on varying $K$ parameter on basketball task. Full-action BC achieves score of 1. TL;DR: Regardless of the $K$ parameter LAPO-slots outperform baseline LAPO.
}
\label{fig:ablations}
\end{figure*}

\subsection{Models}

To conduct the experiments, 4 models were trained: \textbf{LAPO}, \textbf{LAPO-clean}, \textbf{LAPO-slots} and \textbf{LAPO-masks}. All models were trained on the same datasets. More details on training can be found in Supplementary Material.

The baseline model is LAPO, which is trained on observations with distractors, following the \citet{schmidt2024learningactactions} procedure (\textbf{LAPO} in the figures). We use it as a baseline to demonstrate the currently existing limitations of latent action pretraining.

We also trained LAPO on clean data without distractors as our upper bound to illustrate the performance gap caused by distractors (\textbf{LAPO-clean} in the figures). 

LAPO-slots and LAPO-masks are the models that follow the object-centric latent action pre-training pipeline described in \Cref{section:oc-lapo} (respectively, \textbf{LAPO-slots} and \textbf{LAPO-masks} in the figures). LAPO-slots uses slot embeddings extracted from VideoSAUR as input to the latent action model, LAPO-masks uses slot masks, treating each mask as an attention map applied to the original frame.

\subsection{Evaluation performance}

To assess the quality of learned latent actions, we train a BC agent $\pi_\text{BC}(a_t \mid z_t)$ on the inferred latent actions $z_t$, then fine-tune it on a small number  of trajectories ($0.1\%-2.5\%$ from the entire dataset) with true action labels. The small size of the action-labeled sample mirrors real-world deployment, where labeling is expensive. The final performance is measured as \textbf{normalized episodic return} for DCS and \textbf{normalized success rate} for DMW, where a score of 1 corresponds to an oracle BC policy (see Supplementary Material) trained on the entire dataset with ground-truth actions.
As shown in \Cref{table:eval-returns}, both LAPO-slots and LAPO-masks strongly outperform the standard LAPO baseline for both the DCS and DMW domains. Notably: In DCS, LAPO-slots improves the gap between LAPO performance with and without distractors by 50\% on average, with even greater gains in complex tasks like `humanoid-walk` (+174\%). In DMW, the improvements are consistent, with LAPO-slots achieving a +50\% improvement on average over the baseline. These results support our hypothesis: object-centric representations help filter out distractors and retain actionable information. \Cref{fig:eval-returns} shows evaluation returns across increasing numbers of fine-tuning trajectories. Both slot-based methods show better generalization from fewer labeled examples.

\subsection{Slot Selection Investigation}
Object-centric models enable interpretable, modular control by attributing dynamics to individual entities, but identifying which slots encode control-relevant information remains challenging.

We show that linear action probes, trained to predict actions from individual slot representations, can automatically identify relevant slots (method in \Cref{section:oc-lapo}). Probe scores align with manual analysis (\Cref{fig:ablations-A,fig:ablations-B}) and strongly correlate with downstream BC performance (\Cref{fig:ablations-D}), validating them as a reliable relevance metric.

Probing reveals that useful information concentrates in either a small set of dominant slots (one slot for $K \leq 8$ or a tightly coupled pair for $K > 8$), with sharply worse scores for all others. Only these top-ranked slots yield non-zero BC performance (\Cref{fig:ablations-C} for $K=4$ and $K=15$ cases); the rest are ineffective. For $K>8$, concatenating the top two slots improves performance, confirming they capture complementary information. Thus, probe score distributions guide both selection and fusion of meaningful slots. 

Additional visualizations are presented in the Supplementary Material.

\subsection{Varying Number of Slots}
The number of slots $K$ is a key hyperparameter in object-centric models like VideoSAUR~\citep{zadaianchuk2023objectcentric}, often set based on scene complexity. While adaptive allocation has been explored~\citep{fan2024adaptive}, most methods use a fixed $K$, making robustness to its choice crucial for real-world deployment. We evaluate LAPO-slots for $K = 2$ to $15$ on DMW tasks. As shown in \Cref{fig:ablations-E}, performance remains consistently strong across all $K$, demonstrating robustness to the choice of $K$, which is a practical advantage when the number of relevant entities is unknown. Too few slots ($K=2$) cause collapsed representations, merging multiple entities into one. In contrast, larger $K$ introduces duplicated slots with nearly identical representations. For $K=5$, only two unique slots emerge; for $K=8$, four; for $K=11$ or $15$, seven. Despite redundancy, downstream performance remains stable, as slot selection focuses on the most informative ones.

Interestingly, when applied to DMW, we expected a clear separation of distinct objects such as the robotic arm, hammer, ball, etc. Instead, VideoSAUR grouped these task-relevant entities into a single slot even when the number of slots $K$ was greater than the amount of objects.
We attribute this to limited trajectory diversity: variations arise from the manipulated object’s initial position, with little change in arm or target configuration. Without sufficient variability, the model lacks incentive to disentangle co-moving entities. This reflects a general tendency in object-centric models: they favor functional over geometric decomposition, merging semantically distinct but behaviorally coupled entities. Importantly, this behavior mirrors real-world conditions, where data is often imbalanced and incomplete.

\subsection{Increasing distractions difficulty}

To better understand how different types of visual distractions impact latent action learning, we evaluate our methods under increasingly challenging conditions in the Distracting Control Suite (DCS) \citep{stone2021distractingcontrolsuite} environment. In this work, we primarily focus on dynamic backgrounds as the core challenge, but we also test combinations of all three distractor types (backgrounds, camera position, and color variations) on DCS (marked DCS-hard on \Cref{fig:mean-performance} and \Cref{table:eval-returns}) to assess robustness.

As shown in \Cref{fig:mean-performance}, all methods suffer performance degradation under additional distractors, for example, LAPO performance drops by x2.7 under background distractions and by x4.4 under additional distractors, compared to the performance without distractors. However, LAPO-slots maintains a consistent relative improvement over baseline (50\% on DCS $\rightarrow$ 52\% on DCS-hard), while LAPO-masks degrades (50\% on DCS $\rightarrow$ 26\% on DCS-hard). This difference probably stems from VideoSAUR’s use of the DINOv2 encoder \cite{oquab2023dinov2}, which captures high-level semantic features and is more robust to appearance changes than the CNN-based encoder in LAPO and LAPO-masks. These findings suggest that slot-based representations not only isolate task-relevant entities but also encode objects using features that generalize better under real-world visual perturbations.

\section{Related work}

Vision-Language-Action (VLA) models like LAPA \citep{ye2024latentactionpretrainingvideos}, Genie \citep{bruce2024genie}, and UniVLA \citep{bu2025univla} have shown impressive capabilities by training on large-scale video data and deploying on real robots. However, scale can make it difficult to isolate specific failure modes. Our work focuses on one such critical failure: the sensitivity of the underlying Latent Action Models to visual distractors. To enable a controlled study of this problem, we use the Latent Action Policy Optimization (LAPO) framework \citep{schmidt2024learningactactions} within simulated environments. While effective in clean settings, LAPO's performance degradation under distraction makes it an ideal baseline to rigorously evaluate robustness improvements.

Our work confronts this distractor challenge by integrating object-centric learning (OCL). OCL has shown significant promise for compositional generation \citep{Akan2025ICLR}, building structured world models \citep{heins2025axiom, barcellona2024dream} and compositional video prediction \citep{lee2025generative}. The application of OCL to latent action learning has been explored by \citet{villarcorrales2025playslotlearninginverselatent}, who use object-centric models for latent action learning in robotic simulations. However, their work, along with most OCL-based world models, does not investigate performance under heavy visual distraction, a key focus of our paper. Furthermore, their reliance on the SAVi architecture \citep{kipf2021conditional} for object discovery proves less robust in complex, distracting environments than the VideoSAUR model \citep{zadaianchuk2023objectcentric} we employ (see Supplementary Material).

Another work that directly addresses latent action learning with distractors is \citet{nikulin2025latent}. While their goal is similar, their method is orthogonal: they propose a reconstruction-free framework that requires a small set of labeled trajectories to provide supervision during pre-training.
Together, these works highlight the growing interest in robust latent action learning under visual distractions. Our contribution uniquely bridges object-centric perception and latent action modeling, emphasizing slot stability, interpretability, and reduced supervision requirements.

\section{Discussion and Limitations} \label{sec:discuss_limitations}

Our results demonstrate that object-centric (OC) representations can significantly mitigate the impact of distractors when learning latent actions from video. By disentangling scenes into meaningful slots, our approach allows latent action models to focus on causal dynamics rather than spurious correlations, providing a strong inductive bias for learning in noisy environments.

The effectiveness of our method is inherently linked to the capabilities of the underlying OCL model. We believe that as OCL methods continue to improve and scale from simulated to more complex, real-world data, much as VideoSAUR improved upon previous models, so too will the performance and applicability of our approach. Our contribution is to show that integrating object decomposition is a viable path for robust latent action learning, thereby motivating further investment in the underlying OCL technologies.

The reliance on current OCL frameworks introduces several limitations. While powerful segmentation models could isolate objects, they are not inherently designed to handle the dynamic, action-correlated distractors (e.g., camera motion, color shifts) which are explored in our work. The challenges become even greater in scenarios with heavy occlusions or multiple camera viewpoints. Furthermore, current slot-based models like VideoSAUR lack memory mechanisms to handle objects entering or leaving the scene, and often assume a fixed number of objects K. While recent work has explored solutions like dictionary-based architectures and adaptive slot allocation \citep{djukicocebo, fan2024adaptive}, these remain open challenges. 

Finally, we observed that the unsupervised nature of OCL models, particularly under limited data diversity, can lead to non-intuitive collapsed decompositions due to lack of direct control over the learned representations. A promising direction to mitigate this is to augment training data with generative approaches \citep{luo2025solving, sikchi2024rl} or develop weakly-supervised OCL methods to learn more controllable representations.

\section*{Acknowledgments}

This work was supported by the The Ministry of Economic Development of the Russian Federation in accordance with the subsidy agreement (agreement identifier 000000C313925P4H0002; grant No 139-15-2025-012).

\bibliography{main}

\onecolumn
\appendix
\section*{Supplementary Material}
\bigskip
\bigskip

\section{Dataset collection} \label{appendix:dataset-collection}
\bigskip
\bigskip
In this section, we provide further details on the data collection process for the two benchmark environments used in our experiments: the Distracting Control Suite (DCS) and the Distracting Meta-World (DMW). For each environment, we describe how the expert demonstration datasets were generated and outline their key properties.

\textbf{Distracting Control Suite (DCS)}: The datasets were collected via expert policies trained on DCS via PPO (for cheetah-run, walker-run and hopper-hop) and SAC (for humanoid-walk). The scores of the experts are presented on the \Cref{table:expert-bc-scores}. The final dataset of transitions for each task consists of 5k trajectories (1k transitions each). The observations in the dataset have a height and width of 64px. The same underlying expert trajectories were used for both the standard DCS (dynamic backgrounds only) and the \emph{DCS-Hard} settings; the only difference is the visual distractions applied to the observations. The \emph{DCS-Hard} setting combines all three available distractions: dynamic backgrounds, random color variations, and camera perturbations.

\textbf{Distracting MetaWorld (DMW)}: The datasets were collected for the hammer, bin-picking, basketball, and soccer tasks using expert policies from MetaWorld. The scores of the experts are presented on the \Cref{table:expert-bc-scores}. The final dataset of transitions for each task consists of 20k successful trajectories (we filtered out non-successful trajectories during dataset collection). The observations in the dataset have a height and width of 128px.

\begin{table}[h!]
\centering
\begin{tabular}{l|c|c|c|c|c}
\toprule
\bf Task & \bf Expert policy & \bf Expert dataset & \bf BC-vanilla &\ \bf BC  &\ \bf Size, GB 
\\ 
\midrule
cheetah-run & 838 &  838 & 840 & 823 & 58\\
walker-run & 740 & 740 & 735 & 749 & 58\\
hopper-hop & 307 & 307 & 300 & 253 & 58\\
humanoid-walk & 617 & 617 & 601 & 428 & 59\\
\midrule
hammer & 1.0& 1.0 & 1.0 & 1.0 & 61\\
bin-picking & 1.0 & 1.0& 1.0 & 1.0 & 114\\
basketball & 0.96 & 1.0 & 1.0 & 1.0 & 87\\
soccer & 0.88 & 1.0& 1.0 & 0.8 & 67\\

\bottomrule
\end{tabular}
\caption{
Comparison of the Performance of Algorithms. \emph{Expert} denotes  the policy used to collect the dataset trained on privileged information about minimal state of the observation (DCS) and expert policy from (DMW). \emph{BC-vanilla} denotes the scores of behavior cloning agents (BC) trained on full expert dataset to imitate expert policy on the privileged for our method true action labels and non-distracted observations. \emph{BC} denotes the scores of BC agents trained on full expert dataset to imitate expert policy on the distracted observations and privileged for our method true action labels.}
\label{table:expert-bc-scores}
\end{table}


\bigskip
\section{Training details} \label{appendix:training-details}
\bigskip
\bigskip

All experiments were run in H100 GPU 80GB. The models are trained in bfloat16 precision. Training duration is shown in \Cref{table:training-duration}. We perform hyperparameter optimization for each model. The 3 best sets of parameters were evaluated and the best mean score across 3 seeds is reported in \Cref{fig:eval-returns}.

\begin{table}[h!]
\centering
\begin{tabular}{l|r|r}
\toprule
\bf Method & \bf DCS& \bf DMW
\\ 
\midrule
ocp (total) & 92,149,776& 92,149,776  \\
ocp (trainable)&  6,343,440 &  6,343,440 \\
\midrule
lapo-slots & 89,186,432 & 15,491,776\\
lapo-masks & 211,847,849& 36,269,129\\
lapo & 211,847,849  & 36,269,129\\
\midrule
bc & 107,541,504 & 126,703,872\\
\bottomrule
\end{tabular}
\caption{Amount of parameters for different models. The rows represent the following approaches:
\emph{ocp}: denotes the number of parameters of the object-centric model;
\emph{lapo-slots}: denotes the number of parameters for latent action learning from vector representations; utilizing object-centric representations from a precollected dataset;
\emph{lapo}: denotes the number of parameters for latent action learning from images;
\emph{lapo-masks}: denotes the number of parameters for latent action learning from images, utilizing object-centric masks from a precollected dataset (using the same model as \emph{lapo});
\emph{bc}: Denotes the number of parameters of the behavior cloning agent trained on latent actions.
}
\label{table:training-duration}
\end{table}


\begin{table}[h!]
\centering
\begin{tabular}{l|r|r}
\toprule
\bf Method & \bf  DCS & \bf  DMW\\ 
\midrule
ocp        & $\sim$ 6 h 30 m  & $\sim$ 8 h 30 m  \\
lapo-slots & $\sim$ 2 h 30 m& $\sim$ 1 h 30 m  \\
lapo-masks & $\sim$ 7 h 30 m& $\sim$ 2 h 30 m  \\
\midrule
lapo             & $\sim$ 7 h 30 m & $\sim$ 2 h 30 m \\
ocp + lapo-slots & $\sim$ 9 h 0 m  & $\sim$ 10 h 0 m \\
ocp + lapo-masks & $\sim$ 14 h 0 m& $\sim$ 11 h 00 m \\
\midrule
bc + finetuning & $\sim$ 3 h 2 m & $\sim$ 1 h 30 m \\
\bottomrule
\end{tabular}
\caption{Average training duration of the methods. The row represent the following approaches: \emph{ocp} denotes the time spent on object-centric pretraining stage which is common for both slots and masks; \emph{lapo-masks} and \emph{lapo-slots} denote the time spent on training latent action model, reading object-centric representations from a precollected dataset; \emph{lapo} denote the time spent on training latent action model, \emph{ocp + lapo-masks} and \emph{ocp + lapo-slots} denote the time for full pipeline of object-centric latent action learning}
\label{table:training-duration}
\end{table}

Object-centric learning pretraining codebase was adopted from Videosaur \citet{zadaianchuk2023objectcentric}. It utilizes DINOv2 \citep{oquab2023dinov2} pretrained encoder \texttt{vit-base-patch14-dinov2.lvd142m} from Timm \citep{rw2019timm} models hub. The images in the dataset are upscaled for dino encoder up to 518px. The hyperparameters for object-centric pretraining can be found in \Cref{table:hyper-ocl}

Latent action learning model for images and object-centric masks is formed from IDM and FDM models based on a ResNet-like CNN encoder and decoder. The hyperparameter optimization setups for latent action learning from images (used for lapo, lapo-masks) can be found in \Cref{table:hyper-lapo-masks}.

Latent action learning model for object-centric slots is formed from IDM and FDM based on 3-layer MLP blocks with residual connections and GeLU activations to effectively process vector representations. The hyperparameter optimizations setups for latent action learning from representations (used for lapo-slots) can be found in \Cref{table:hyper-lapo-masks}.

\bigskip
\bigskip
\section{Fixed Initialization for Slot Stability} \label{appendix:fixed-init}
\bigskip
To mitigate slot permutation variance across predictions, we introduce a fixed slot initialization scheme that learns deterministic initial slot vectors while preserving robustness. Unlike standard Gaussian initialization, which samples slots stochastically at each step, our approach learns per-slot parameters (mean $\boldsymbol{\mu}_k \in \mathbb{R}^d$ and variance $\boldsymbol{\sigma}_k \in \mathbb{R}^d$) during training. During training, we inject controlled noise scaled by the learned variance into the slot initializations, acting as a regularizer to encourage robust feature disentanglement. At inference, slots are initialized deterministically using the learned means, ensuring consistent slot-object assignments. This hybrid strategy bridges the gap between training stability and inference consistency: the noise-augmented training phase prevents overfitting to fixed initializations, while the deterministic inference phase enables efficient object-wise slot selection via decoder masks as visual priors. 
\bigskip
$$
\texttt{Train:} \quad \boldsymbol{s}_k^{\text{(init)}} = \boldsymbol{\mu}_k + \boldsymbol{\sigma}_k \odot \boldsymbol{\epsilon}, \quad \boldsymbol{\epsilon} \sim \mathcal{N}(\boldsymbol{0}, \boldsymbol{I}), \quad \texttt{Inference:} \quad
\boldsymbol{s}_k^{\text{(inference)}} = \boldsymbol{\mu}_k.
$$
\bigskip
\bigskip

\newpage
\section{Evaluation scores on DCS-hard}
\bigskip

We mark as \emph{DCS-hard}, the experiments on Disctracting Control Suite with 3 types of distractions: dynamic backgroungs, color and camera position perturbations. The scale of color and camera variations is $0.1$. The aggregated mean scores for \emph{DCS-hard} and the corresponding results on \emph{DCS} (only dynamic backgrounds) are present in the Main Paper.

\begin{figure}[h!]
    \vskip 0.2in
    \begin{center}
        \centerline{\includegraphics[width=\textwidth]{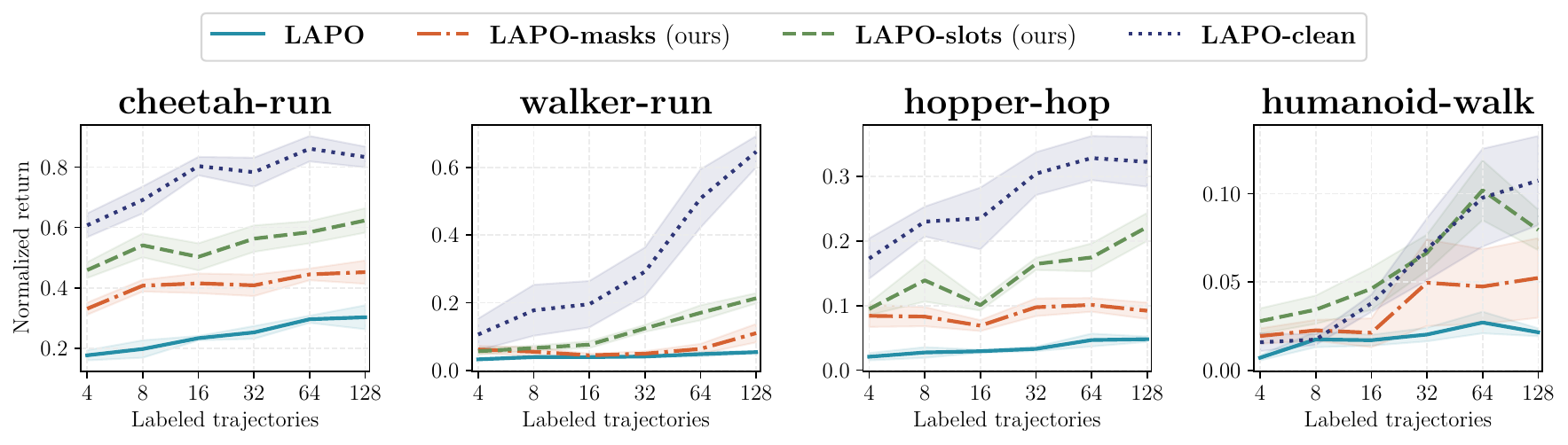}}
        \caption{
        Normalized evaluation returns of the BC agent trained on latent actions for varying numbers of fine-tuning labeled trajectories. 
        TL;DR: Object-centric learning improves evaluation returns in DCS tasks with all 3 types of distractions.
        The plots are arranged from left to right in order of increasing task complexity. The values are averaged across three random seeds.
        The BC agent trained with access to the full dataset of ground-truth actions would return a score of 1 for each task. 
        }
        \label{fig:supp-eval-returns-hard}
    \end{center}
    \vskip -0.2in
\end{figure}

\bigskip
\section{Correlation between probe and normalized success rates} \label{appendix:correlation}
\bigskip
We examine the correlation between the linear action probe and the evaluation performance, using the same budget of labeled trajectories for both probe training and supervised fine-tuning of object-centric latent action learning (LAPO-slots), on the basketball task from DMW.

\begin{figure}[h!]
  \centering
  \def\myheight{4.5cm}
  
  \includegraphics[height=\myheight]{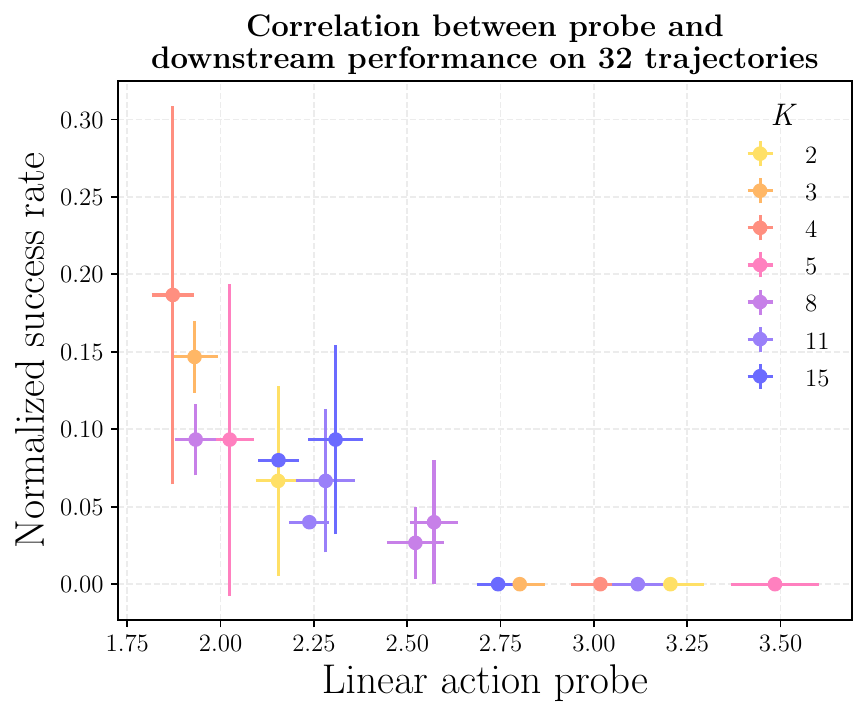}%
  \hfill
  \includegraphics[height=\myheight]{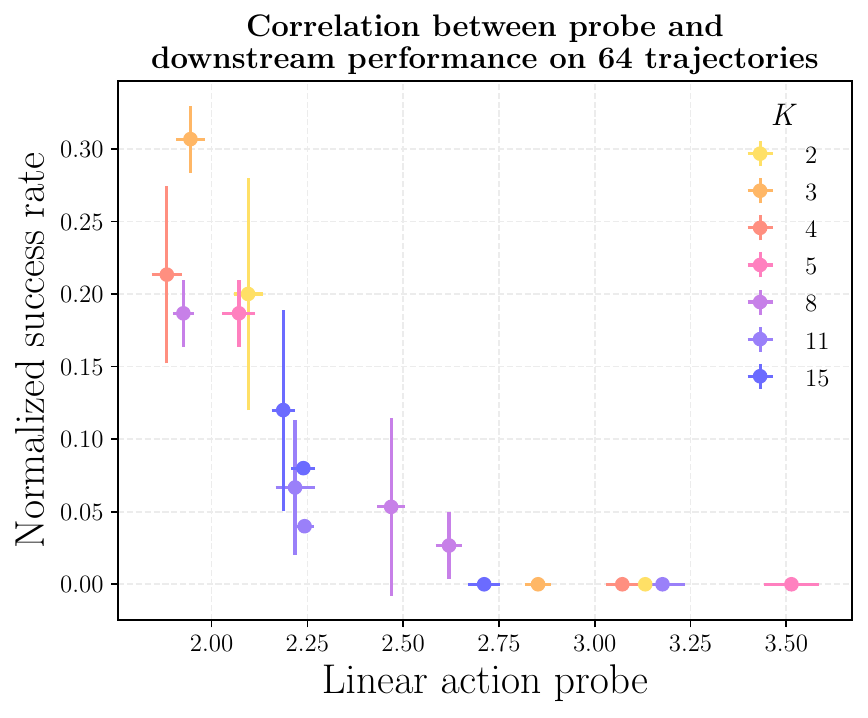}%
  \hfill
  \includegraphics[height=\myheight]{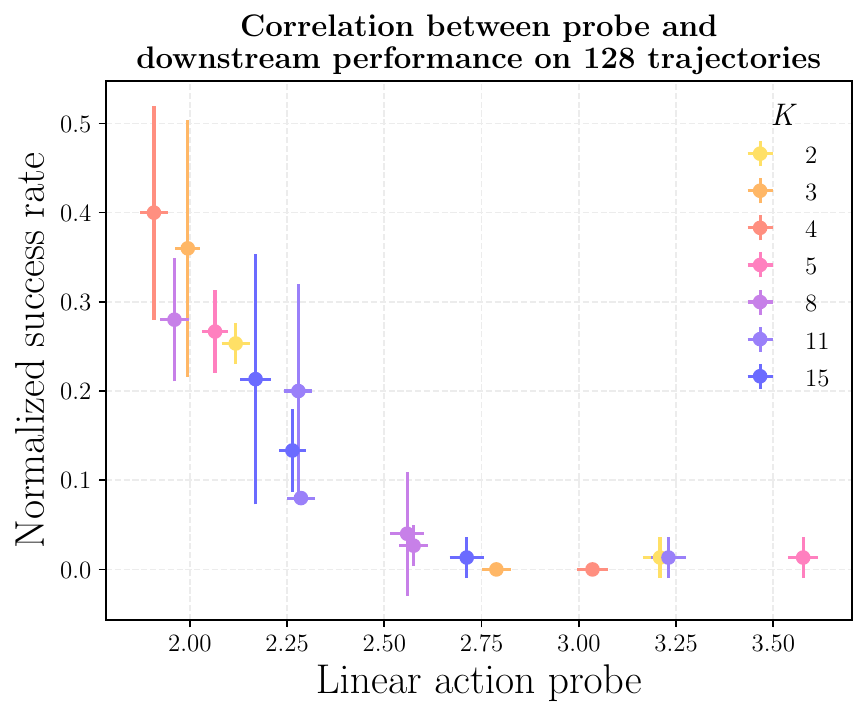}%
  
  \caption{Linear action probes and corresponding resulting normalized success rates of BC agents on same amounts of labeled trajectories for different number of slots (top-2 slots for $K\leq5$, top-3 slots for $K\geq8$). From top to bottom: (1) 32 labeled trajectories, (2) 64 labeled trajectories, (3) 128 labeled trajectories}
  \label{fig:probe-corr-supp}
\end{figure}

As shown in \Cref{fig:probe-corr-supp}, there is a strong correlation between the probe performance and the resulting success rate, suggesting that the probe can serve as a reliable method for slot selection.

\newpage
\section{Probe slots selection for DCS} \label{appendix:probe-DCS}
\bigskip
Slot selection for tasks from the Distracting Control Suite: linear probes on different slots and corresponding visual masks examples. The slots with the lowest linear probe (\textbf{top I}) were used to obtain the results of tables and figures in the Main Paper.
\bigskip

\begin{figure}[h!]
  \centering
  \def\myheight{2.7cm}
  
  \includegraphics[height=\myheight]{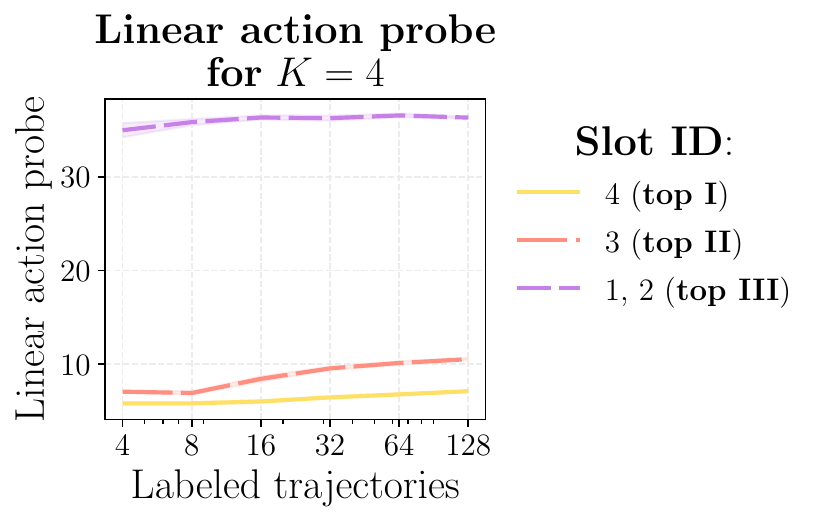}%
  \hfill
  \includegraphics[height=\myheight]{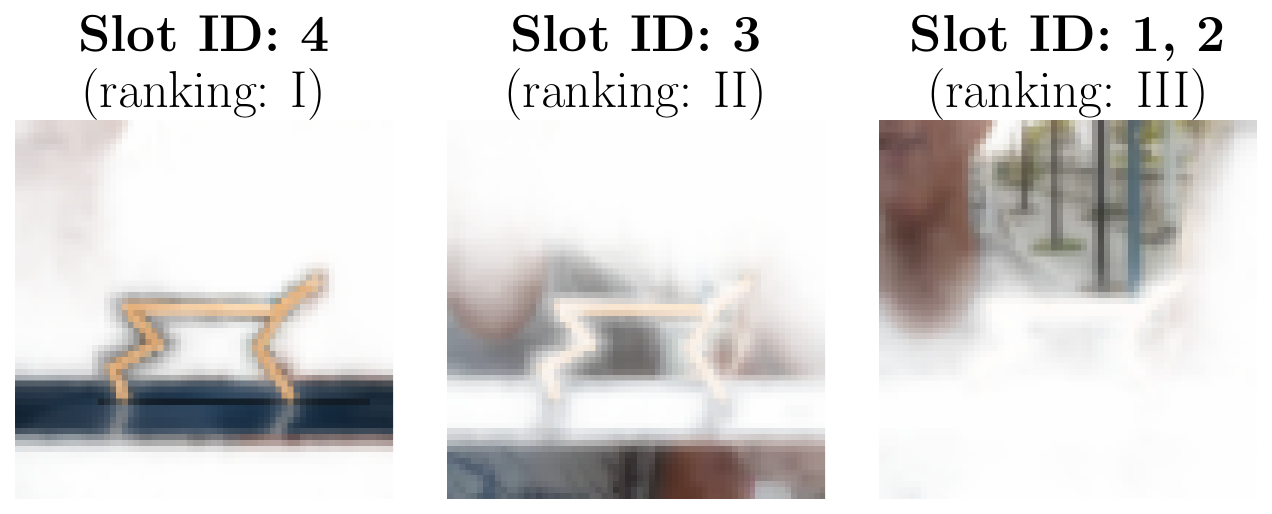}%
  
  \caption{Slot selection for \emph{cheetah-run}. From left to right: (1) Linear action probes for VideoSaur slots pretrained cheetah-run task for varying budget of trajectories. (2) Corresponding slot masks for cheetah-run task after VideoSaur pretraining for number of slots $K=4$. }
  \label{fig:probe-corr-cheetah-4}
\end{figure}

\begin{figure}[h!]
  \centering
  \def\myheight{2.7cm}
  
  \includegraphics[height=\myheight]{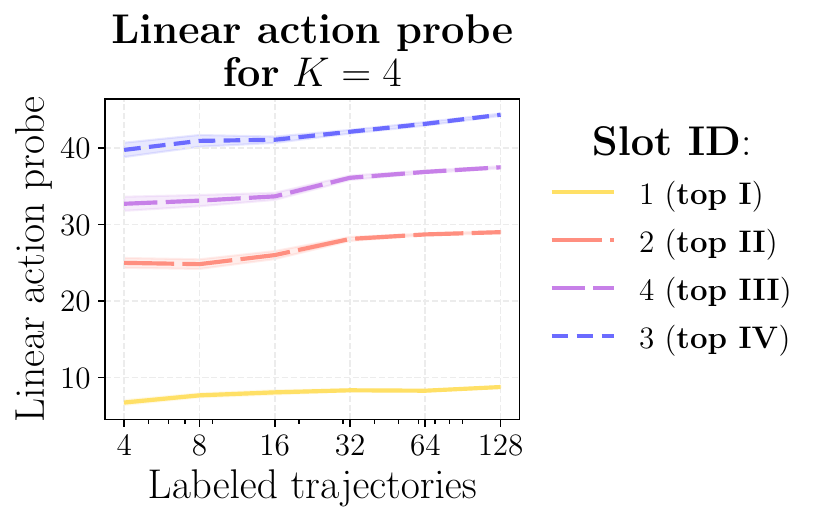}%
  \hfill
  \includegraphics[height=\myheight]{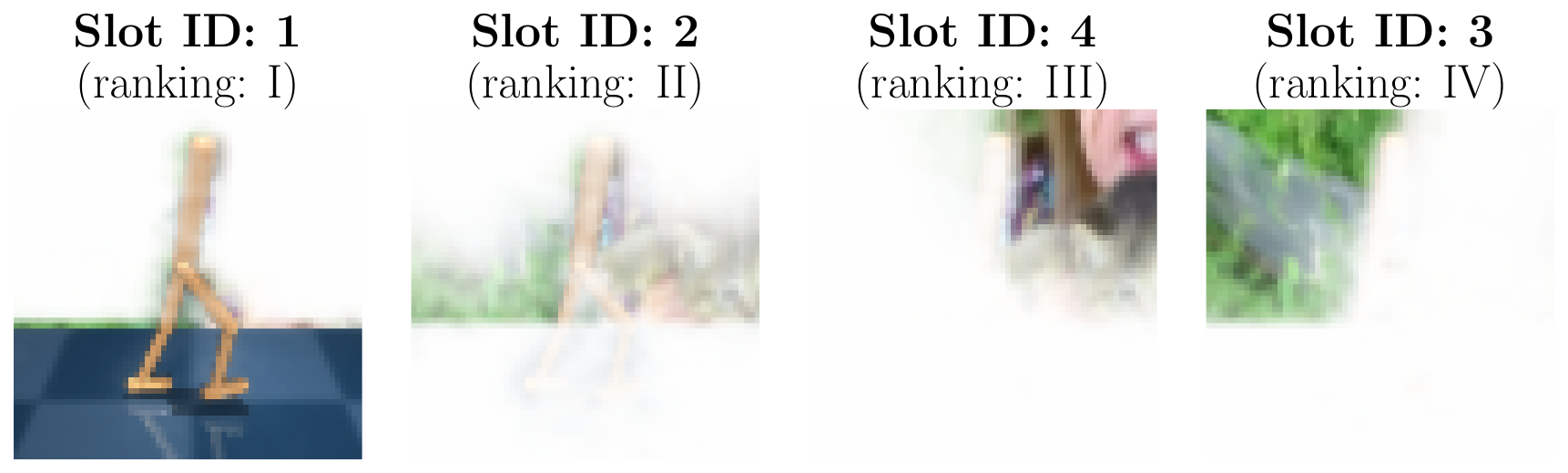}%
  \caption{Slot selection for \emph{walker-run}. From left to right: (1) Linear action probes for VideoSaur slots pretrained walker-run task for varying budget of trajectories. (2) Corresponding slot masks for walker-run task after VideoSaur pretraining for number of slots $K=4$. }
  \label{fig:probe-corr-walker-4}
\end{figure}

\begin{figure}[h!]
  \centering
  \def\myheight{2.7cm}
  \includegraphics[height=\myheight]{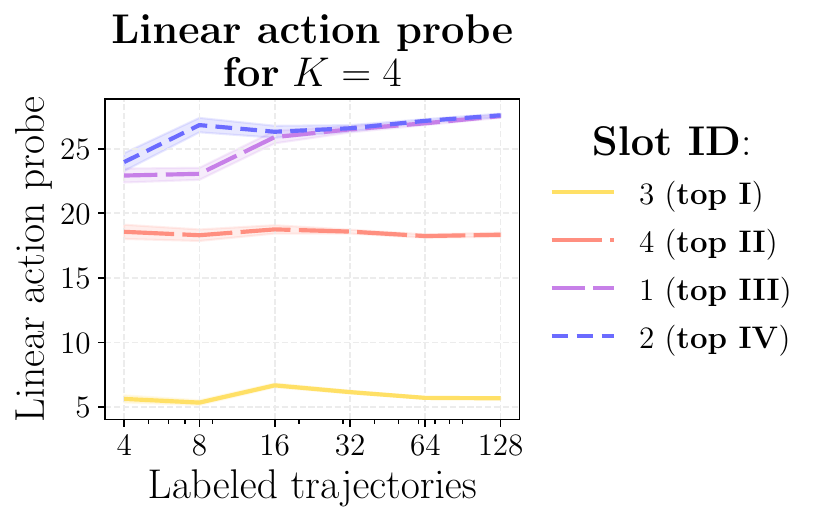}%
  \hfill
  \includegraphics[height=\myheight]{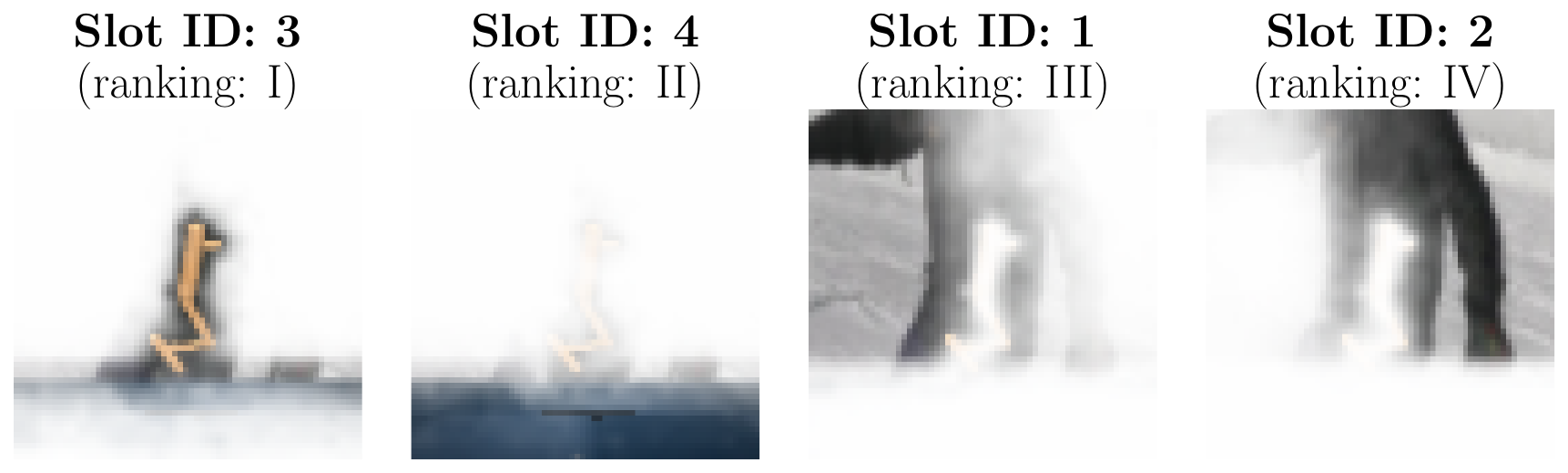}%
  
  \caption{Slot selection for \emph{hopper-hop}. From left to right: (1) Linear action probes for VideoSaur slots pretrained hopper-hop task for varying budget of trajectories. (2) Corresponding slot masks for hopper-hop task after VideoSaur pretraining for number of slots $K=4$. }
  \label{fig:probe-corr-hopper-4}
\end{figure}

\begin{figure}[h!]
  \centering
  \def\myheight{2.7cm}
  
  \includegraphics[height=\myheight]{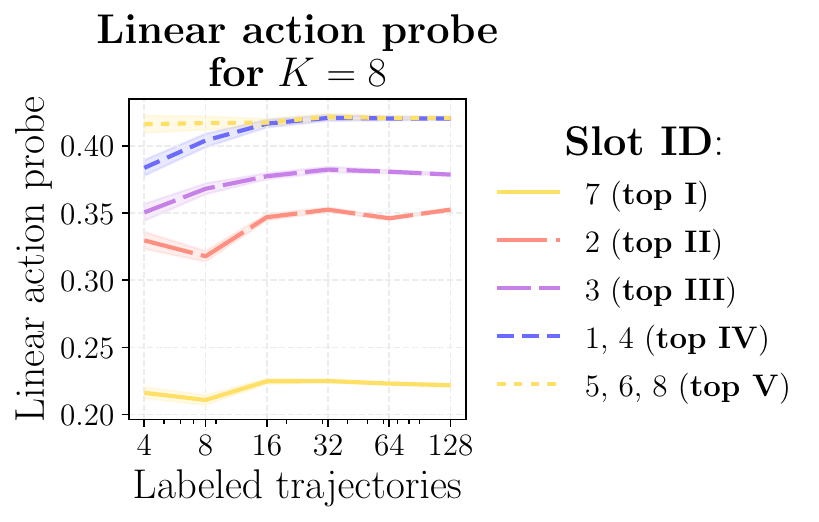}%
  \hfill
  \includegraphics[height=\myheight]{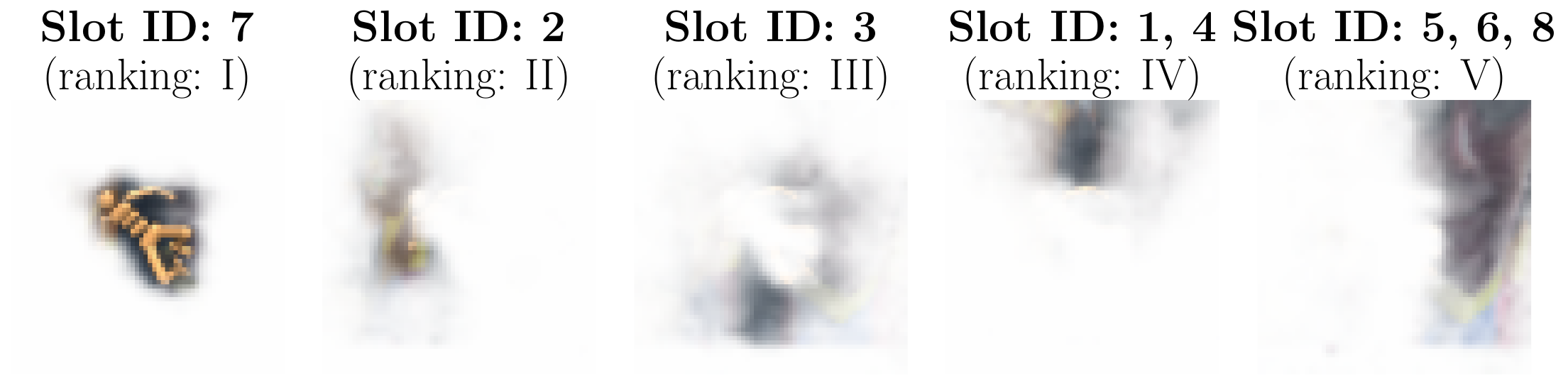}%
  
  \caption{Slot selection for \emph{humanoid-walk}. From left to right: (1) Linear action probes for VideoSaur slots pretrained humanoid-walk task for varying budget of trajectories. (2) Corresponding slot masks for humanoid-walk task after VideoSaur pretraining for number of slots $K=8$.}
  \label{fig:probe-corr-humanoid-8}
\end{figure}


\newpage
\section{Probe slots selection for DMW} \label{appendix:probe-DMW}
\bigskip
Slot selection for tasks from the Distracting MetaWord: linear probes on different slots and corresponding visual masks examples. The slots with the lowest linear probe (\textbf{top I}) were used to obtain the results of tables and figures in the Main Paper.
\bigskip

\begin{figure}[h!]
  \centering
  \def\myheight{2.7cm}
  
  \includegraphics[height=\myheight]{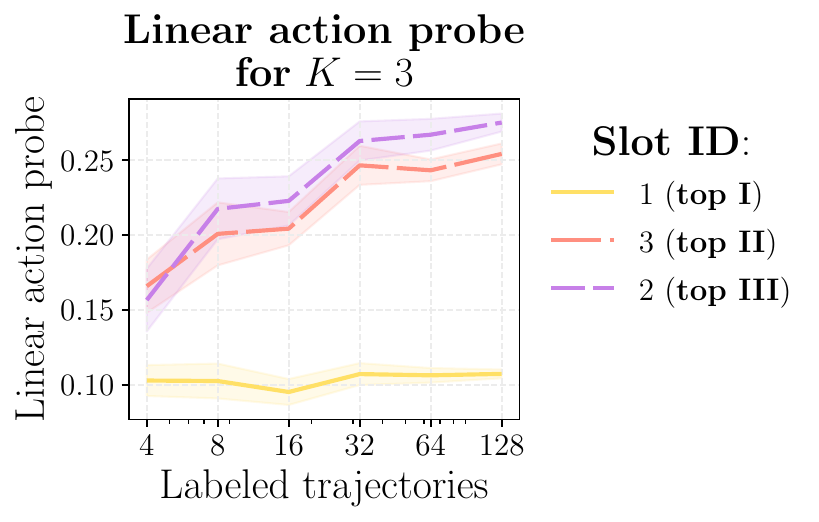}%
  \hfill
  \includegraphics[height=\myheight]{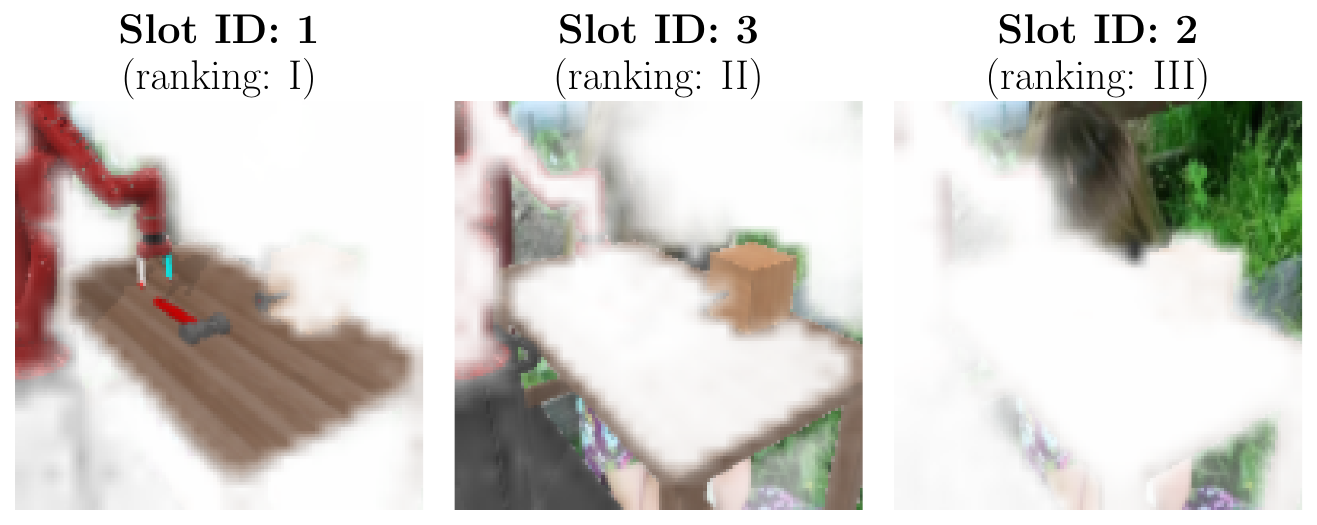}%
  
  \caption{Slot selection for \emph{hammer}. From left to right: (1) Linear action probes for VideoSaur slots pretrained hammer task for varying budget of trajectories. (2) Corresponding slot masks for hammer task after VideoSaur pretraining for number of slots $K=3$. }
  \label{fig:probe-corr-hammer-3}
\end{figure}

\begin{figure}[h!]
  \centering
  \def\myheight{2.7cm}
  
  \includegraphics[height=\myheight]{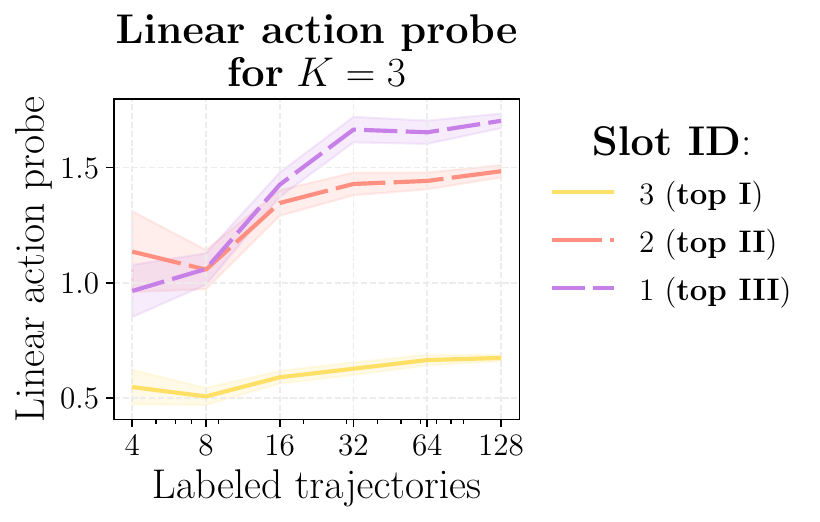}%
  \hfill
  \includegraphics[height=\myheight]{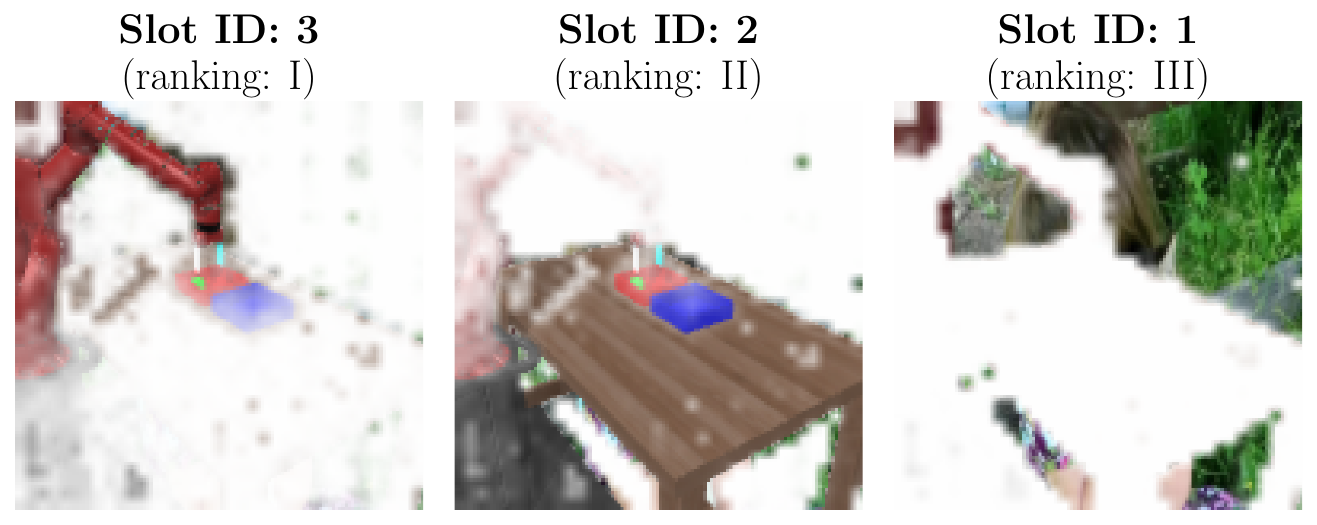}%
  
  \caption{Slot selection for \emph{bin-picking}. From left to right: (1) Linear action probes for VideoSaur slots pretrained bin-picking task for varying budget of trajectories. (2) Corresponding slot masks for bin-picking task after VideoSaur pretraining for number of slots $K=3$. }
  \label{fig:probe-corr-bin-3}
\end{figure}

\begin{figure}[h!]
  \centering
  \def\myheight{2.7cm}

  \includegraphics[height=\myheight]{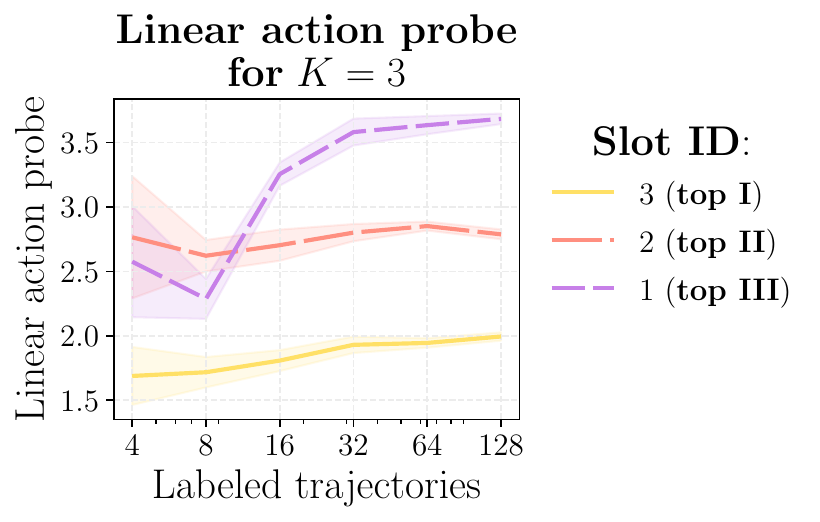}%
  \hfill
  \includegraphics[height=\myheight]{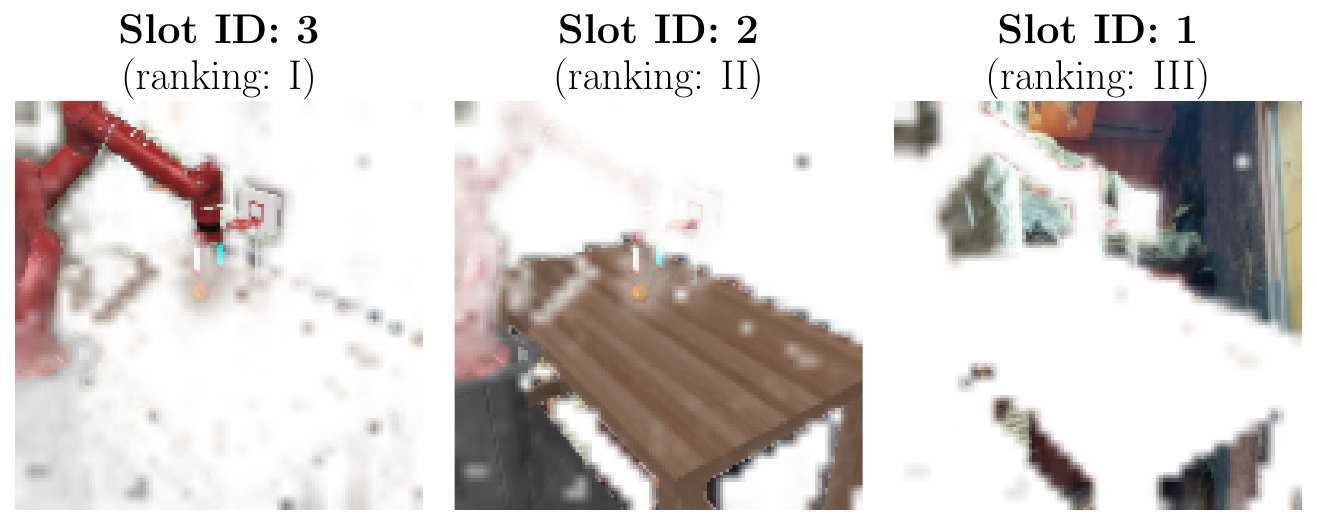}%
  
  \caption{Slot selection for \emph{basketball}. From left to right: (1) Linear action probes for VideoSaur slots pretrained basketball task for varying budget of trajectories. (2) Corresponding slot masks for basketball task after VideoSaur pretraining for number of slots $K=3$.}
  \label{fig:probe-corr-basketball-3}
\end{figure}

\begin{figure}[h!]
  \centering
  \def\myheight{2.7cm}
  
  \includegraphics[height=\myheight]{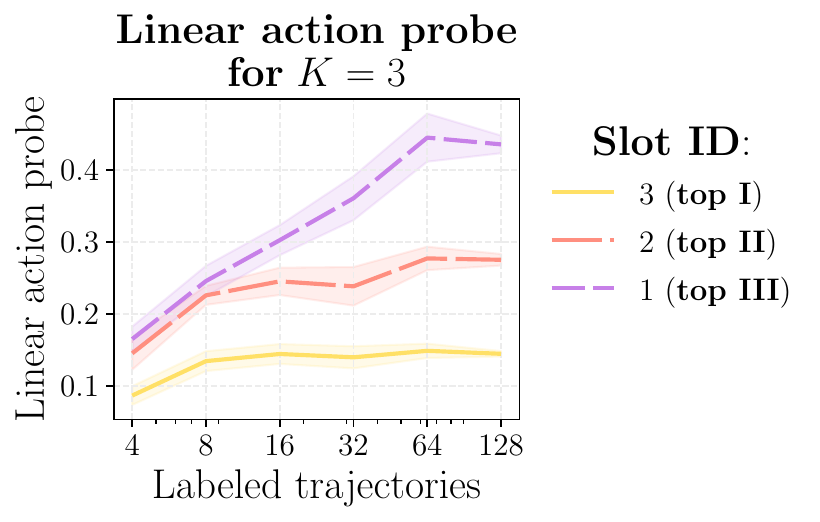}%
  \hfill
  \includegraphics[height=\myheight]{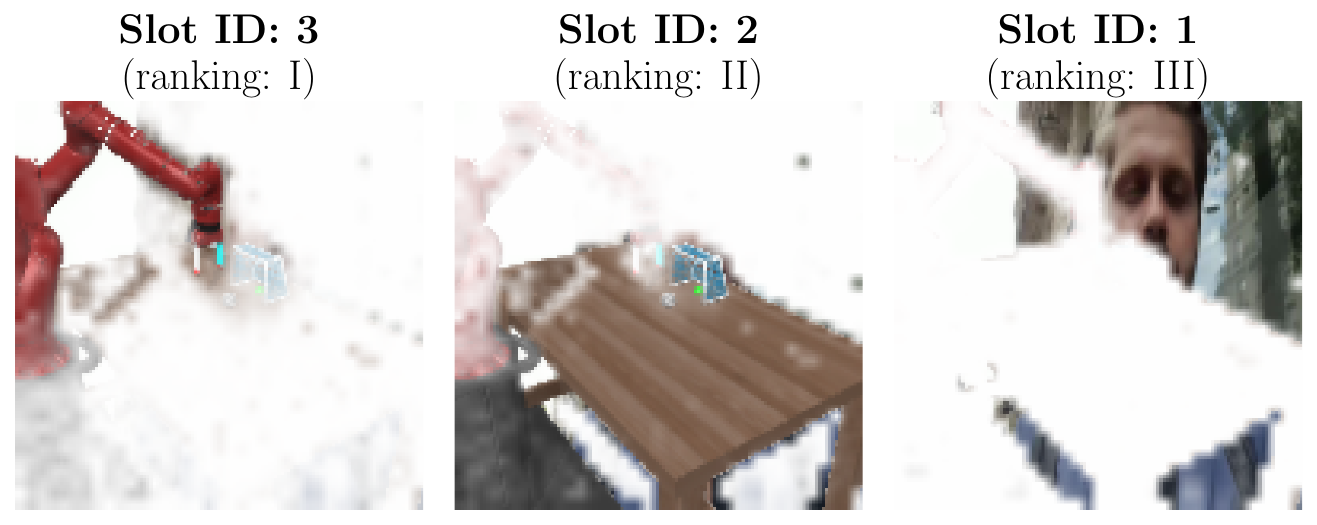}%
  
  \caption{Slot selection for \emph{soccer}. From left to right: (1) Linear action probes for VideoSaur slots pretrained soccer task for varying budget of trajectories. (2) Corresponding slot masks for soccer task after VideoSaur pretraining for number of slots $K=3$. }
  \label{fig:probe-corr-soccer-3}
\end{figure}

\newpage
\section{Probe slots selection for different number of slots} \label{appendix:varying-num-slots}
\bigskip

\begin{figure}[h!]
  \centering
  \def\myheight{2.8cm}
  
  \includegraphics[height=\myheight]{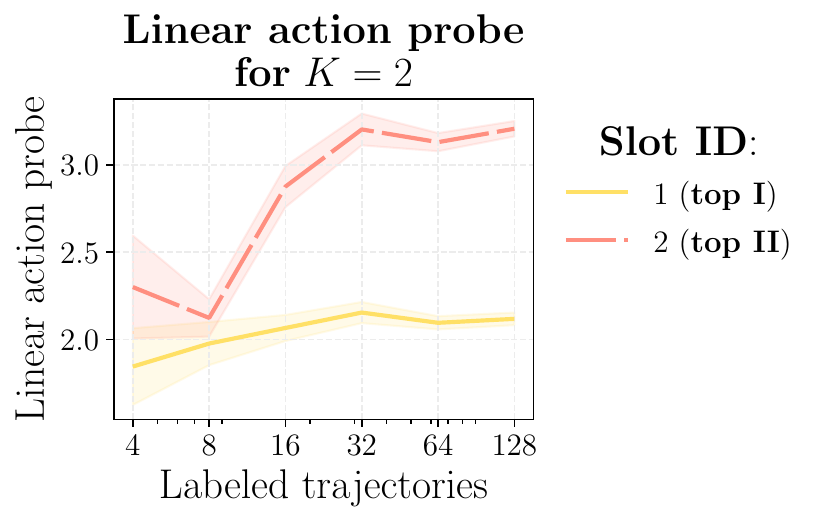}%
  \hfill
  \includegraphics[height=\myheight]{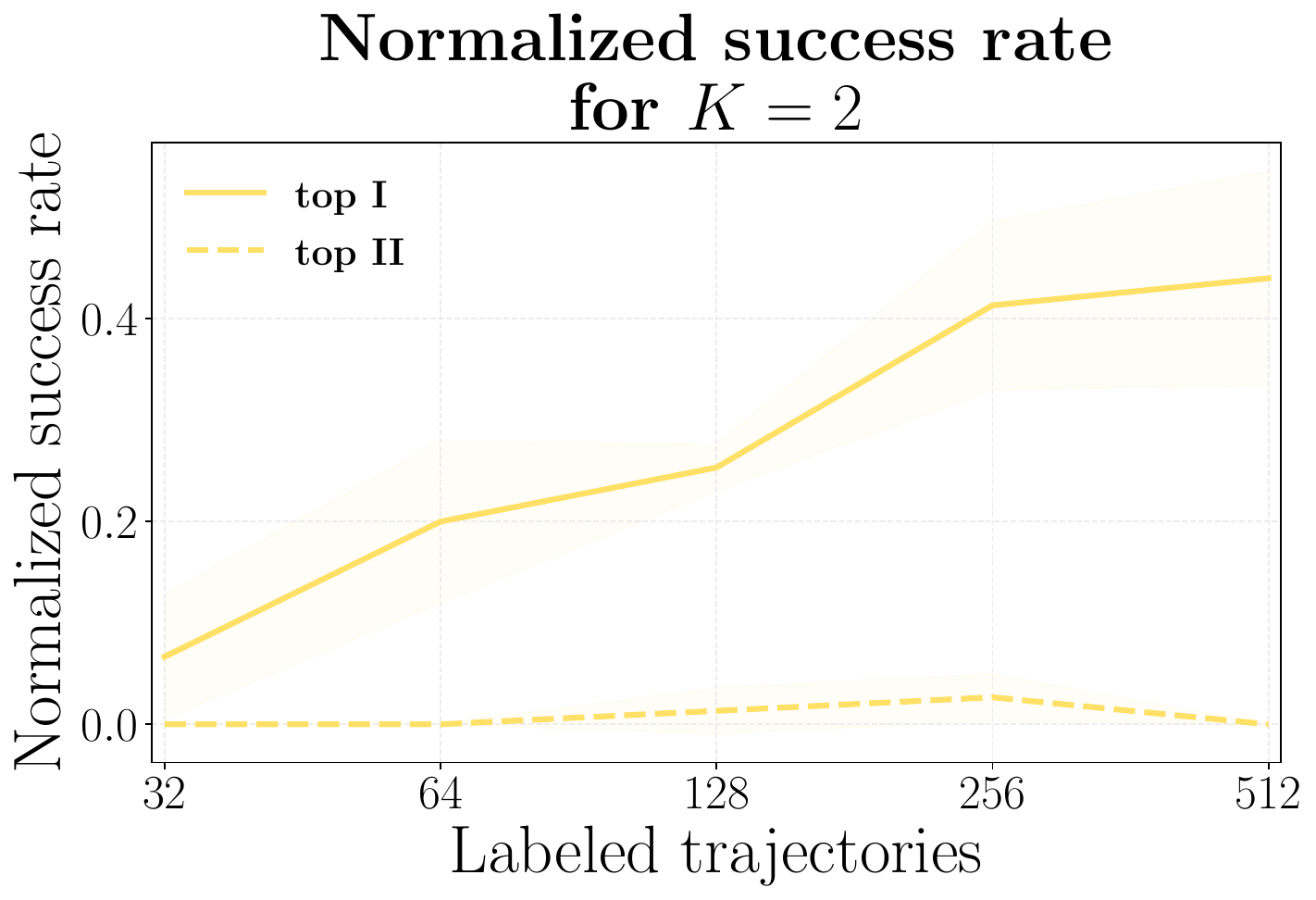}%
  \hfill
  \includegraphics[height=\myheight]{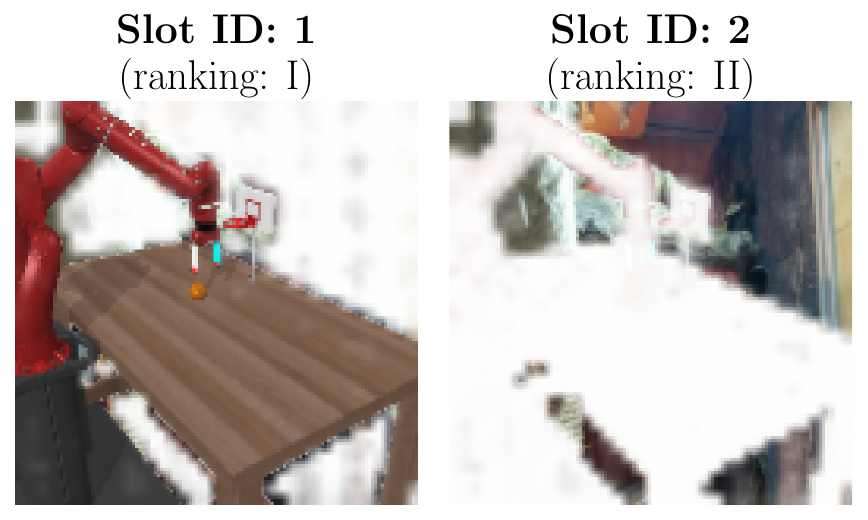}%
  
  \caption{Slot selection for \emph{2} slots. From left to right: (1) Linear action probes for VideoSaur slots pretrained basketball task for varying budget of trajectories. (2) Downstream performance of top I and top II slots pretrained basketball task for varying budget of trajectories. (3) Corresponding slot masks for basketball task after VideoSaur pretraining for number of slots $K=2$.}
  \label{fig:probe-corr-basketball-2}
\end{figure}

Slot selection for varying number of slots for basketball task from Distracted MetaWord: linear probes on different slots, downstream performance and corresponding visual masks examples. The slots with the lowest linear probe (top I for \Cref{fig:probe-corr-basketball-2,fig:probe-corr-basketball-3,fig:probe-corr-basketball-5,fig:probe-corr-basketball-8} and top I+II for \Cref{fig:probe-corr-basketball-11,fig:probe-corr-basketball-15}) were used to obtain the results Figure 5e in the Main Paper.
\bigskip

\begin{figure}[h!]
  \centering
  \def\myheight{2.6cm}
  
  \includegraphics[height=\myheight]{plots/DMW-basketball-3-probes.pdf}%
  \hfill
  \includegraphics[height=\myheight]{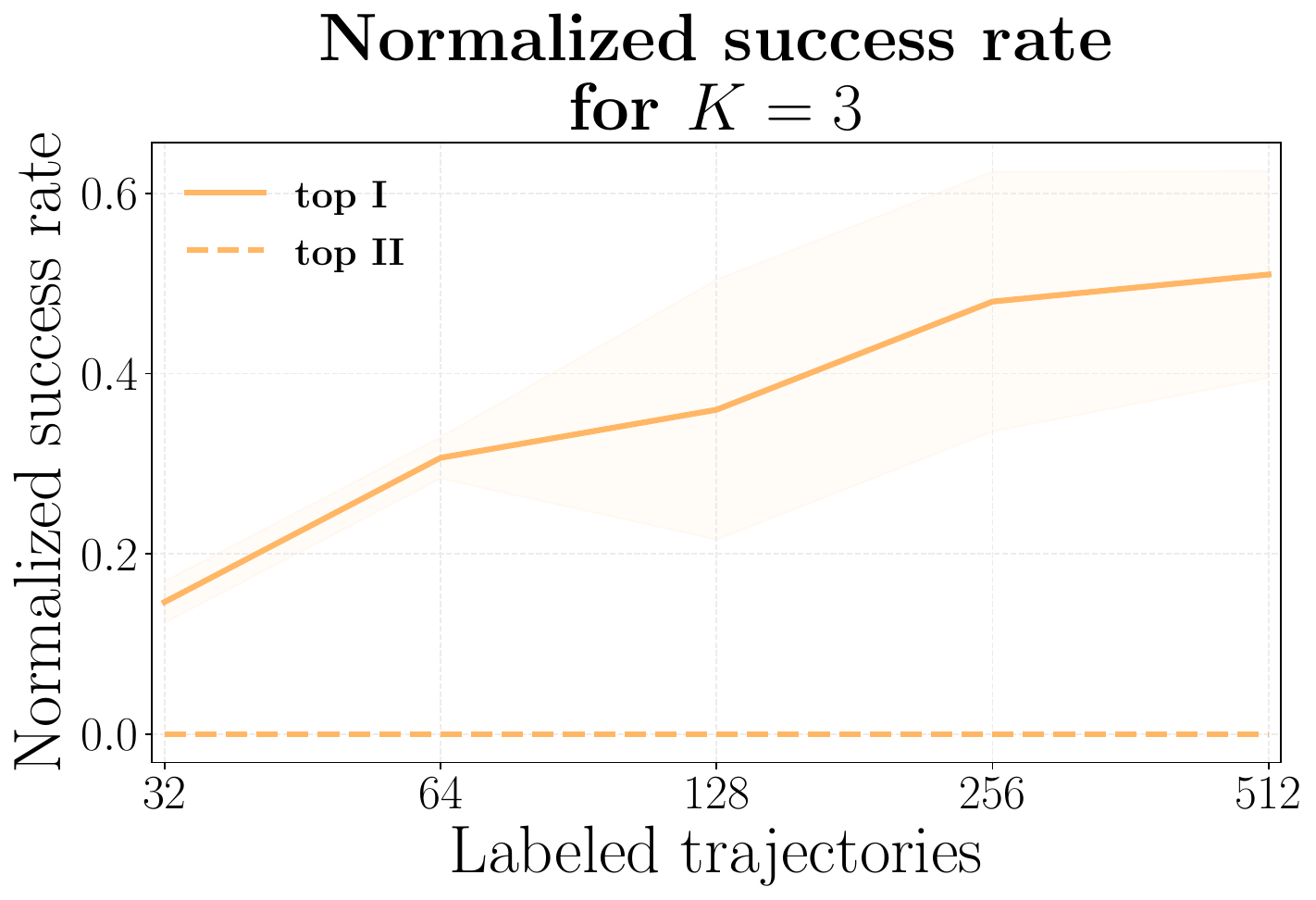}%
  \hfill
  \includegraphics[height=\myheight]{plots/DMW-basketball-3-masks.pdf}%
  
  \caption{Slot selection for \emph{3} slots. From left to right: (1) Linear action probes for VideoSaur slots pretrained basketball task for varying budget of trajectories. (2) Downstream performance of top I and top II slots pretrained basketball task for varying budget of trajectories. (3) Corresponding slot masks for basketball task after VideoSaur pretraining for number of slots $K=3$.}
  \label{fig:probe-corr-basketball-3}
\end{figure}

\begin{figure}[h!]
  \centering
  \def\myheight{3.2cm}
  
  \includegraphics[height=\myheight]{plots/DMW-basketball-4-probes.pdf}%
  \hfill
  \includegraphics[height=\myheight]{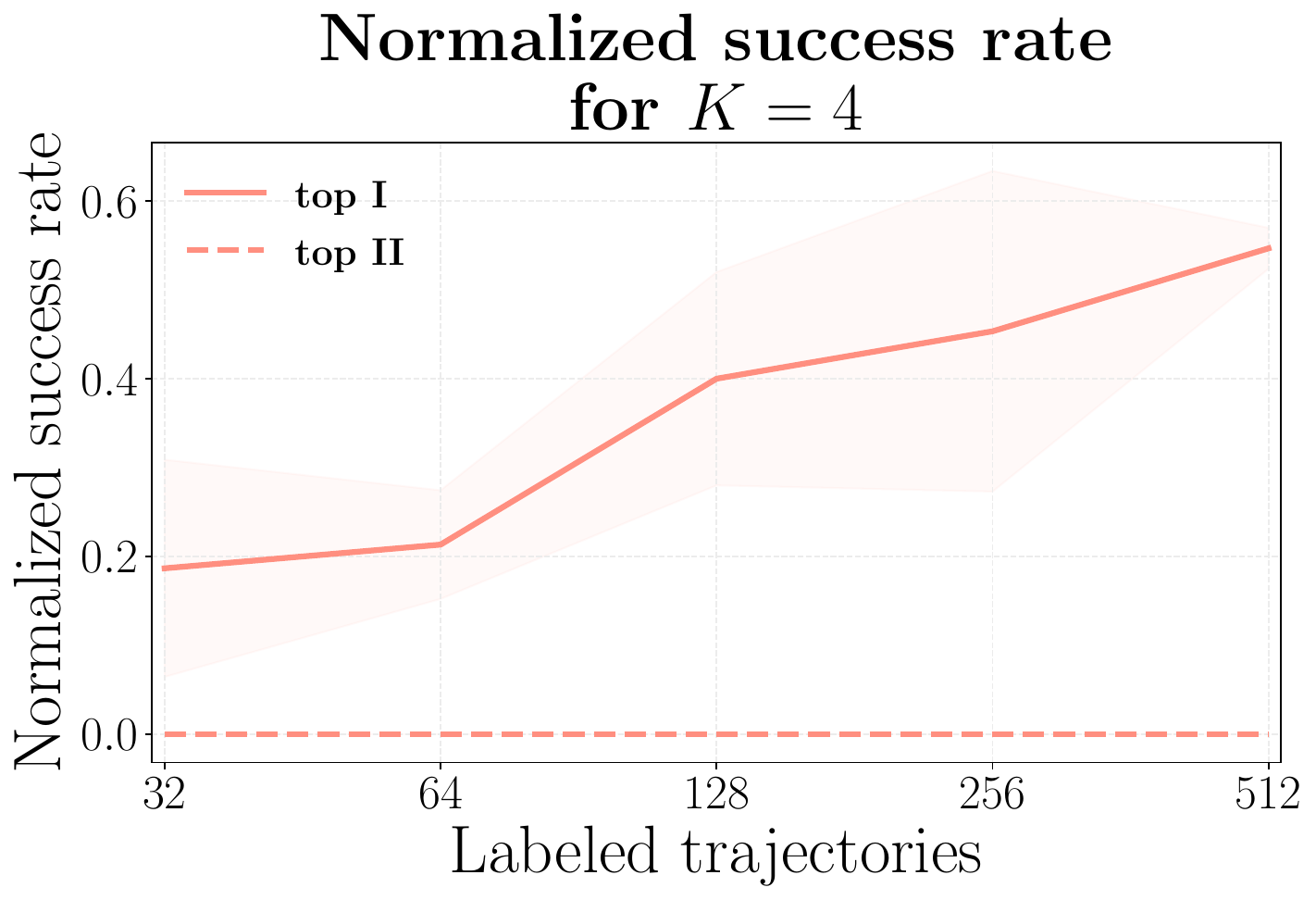}%
  \hfill
  \includegraphics[height=\myheight]{plots/DMW-basketball-4-masks.pdf}%
  
  \caption{Slot selection for \emph{4} slots. From left to right: (1) Linear action probes for VideoSaur slots pretrained basketball task for varying budget of trajectories. (2) Downstream performance of top I and top II slots pretrained basketball task for varying budget of trajectories. (3) Corresponding slot masks for basketball task after VideoSaur pretraining for number of slots $K=4$.}
  \label{fig:probe-corr-basketball-4}
\end{figure}

\begin{figure}[h!]
  \centering
  \def\myheight{2.6cm}
  
  \includegraphics[height=\myheight]{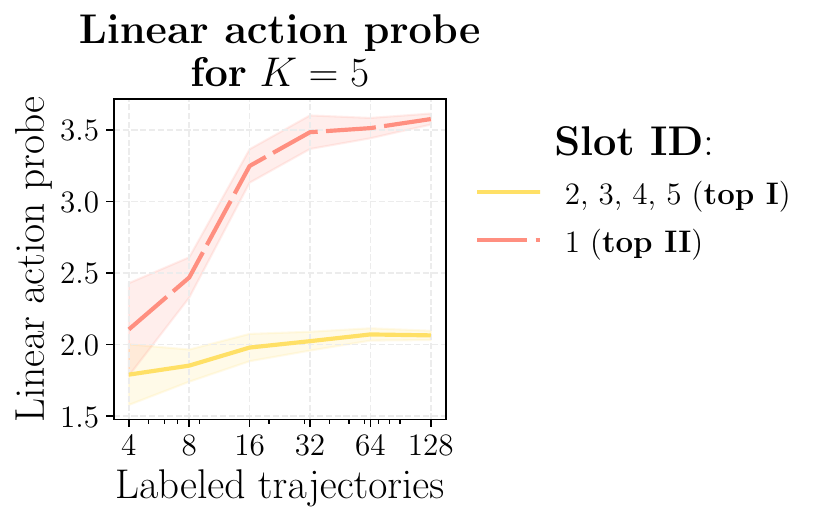}%
  \hfill
  \includegraphics[height=\myheight]{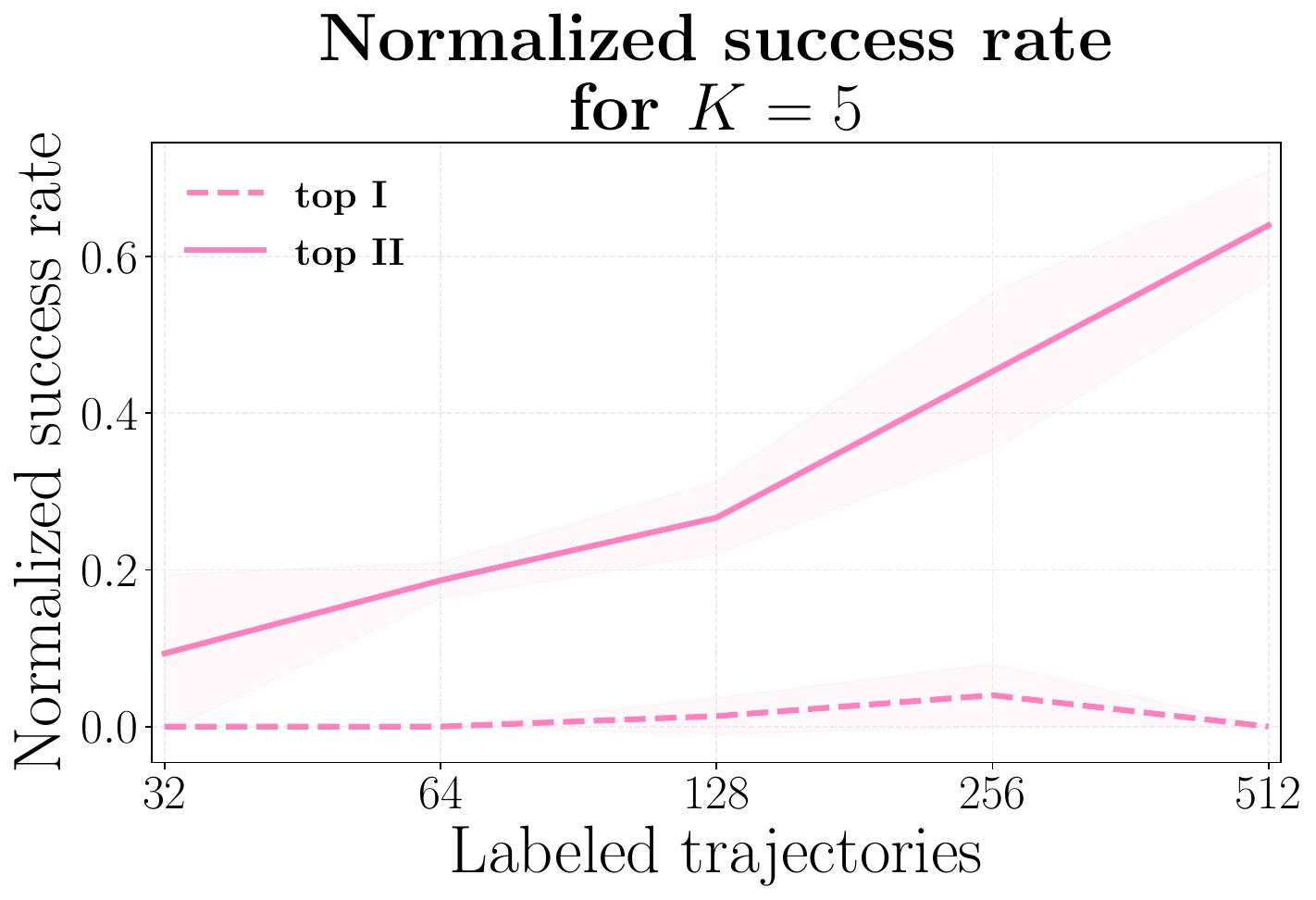}%
  \hfill
  \includegraphics[height=\myheight]{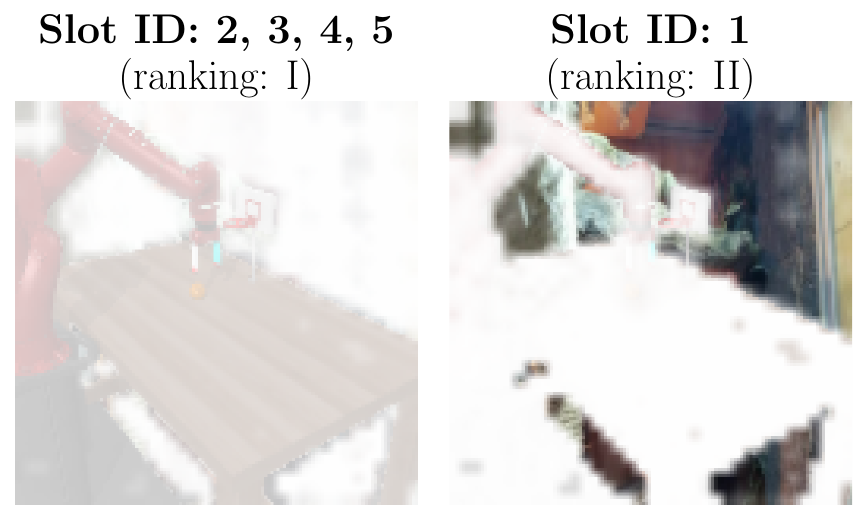}%
  
  \caption{Slot selection for \emph{5} slots. From left to right: (1) Linear action probes for VideoSaur slots pretrained basketball task for varying budget of trajectories. (2) Downstream performance of top I and top II slots pretrained basketball task for varying budget of trajectories. (3) Corresponding slot masks for basketball task after VideoSaur pretraining for number of slots $K=5$.}
  \label{fig:probe-corr-basketball-5}
\end{figure}

\begin{figure}[h!]
  \centering
  \def\myheight{2.6cm}
  
  \includegraphics[height=\myheight]{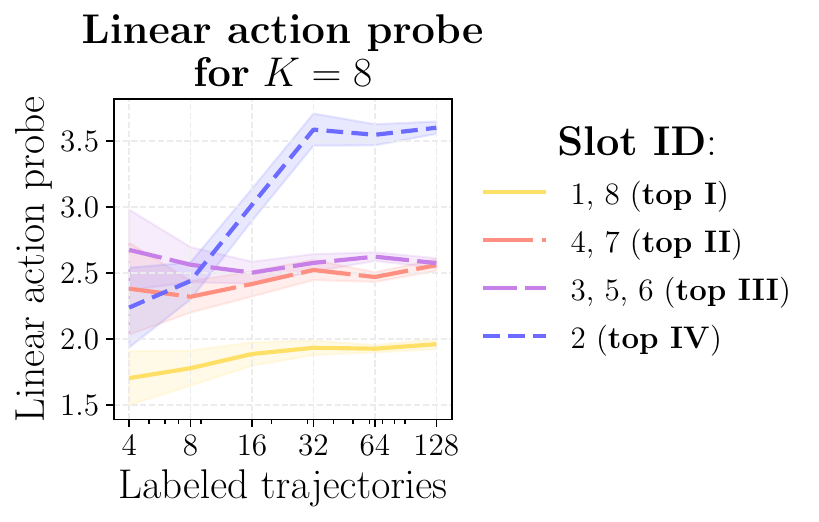}%
  \hfill
  \includegraphics[height=\myheight]{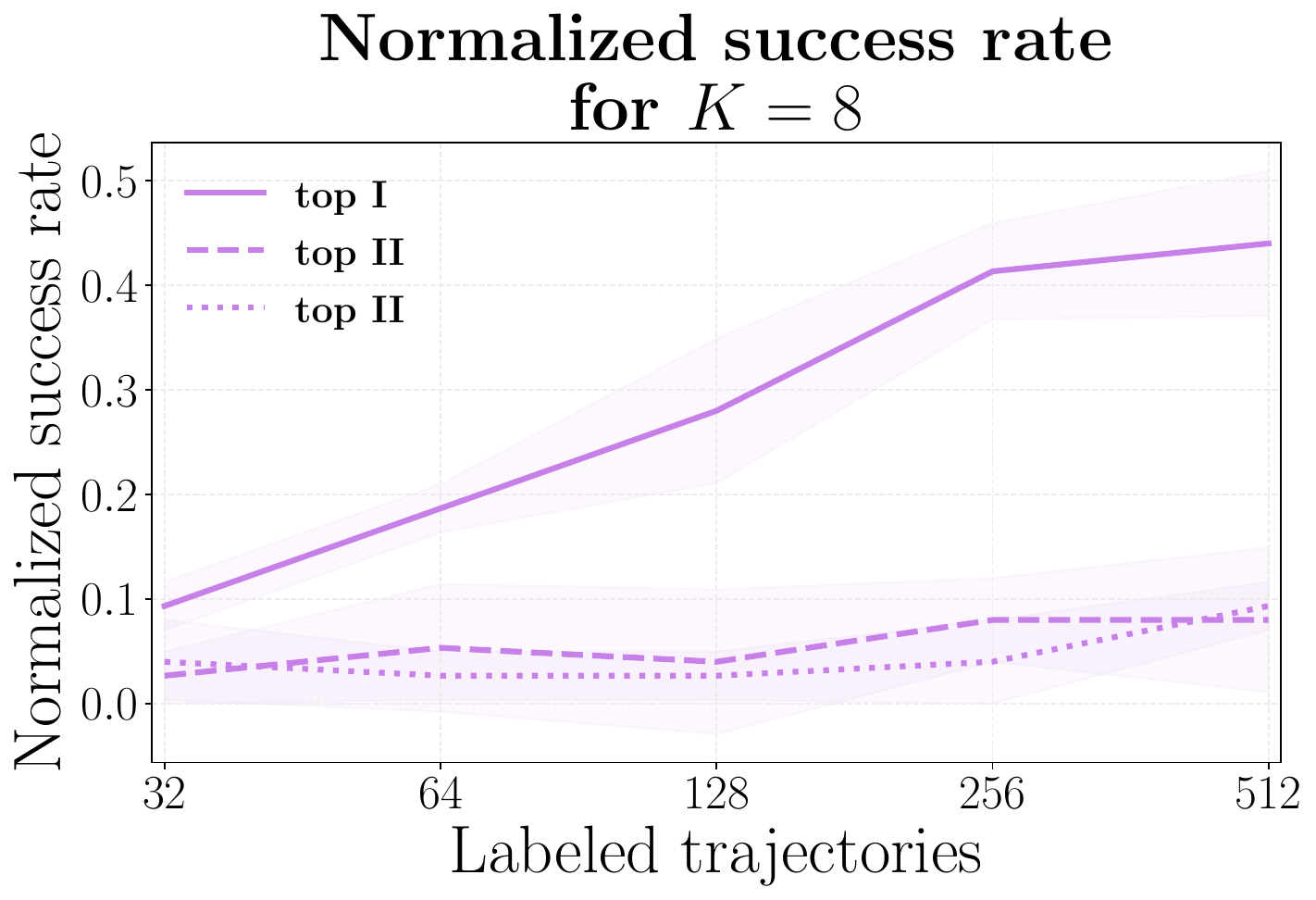}%
  \hfill
  \includegraphics[height=\myheight]{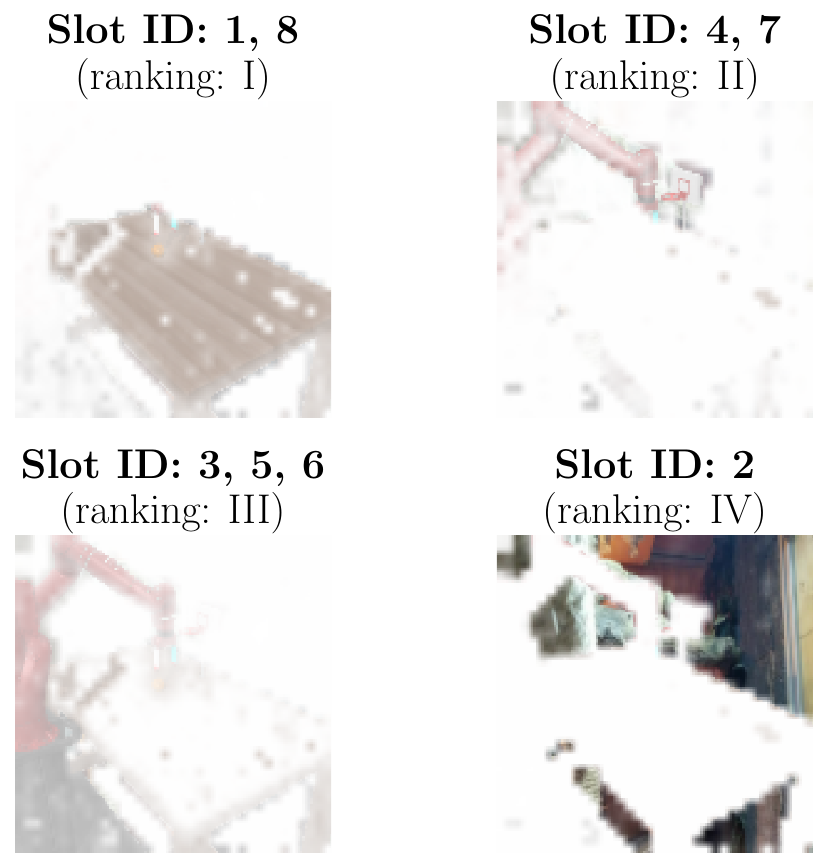}%
  
  \caption{Slot selection for \emph{8} slots. From left to right: (1) Linear action probes for VideoSaur slots pretrained basketball task for varying budget of trajectories. (2) Downstream performance of top I and top II slots pretrained basketball task for varying budget of trajectories. (3) Corresponding slot masks for basketball task after VideoSaur pretraining for number of slots $K=8$.}
  \label{fig:probe-corr-basketball-8}
\end{figure}

\begin{figure}[h!]
  \centering
  \def\myheightt{4cm}
  \def\myheight{2.6cm}
  
  \includegraphics[height=\myheightt]{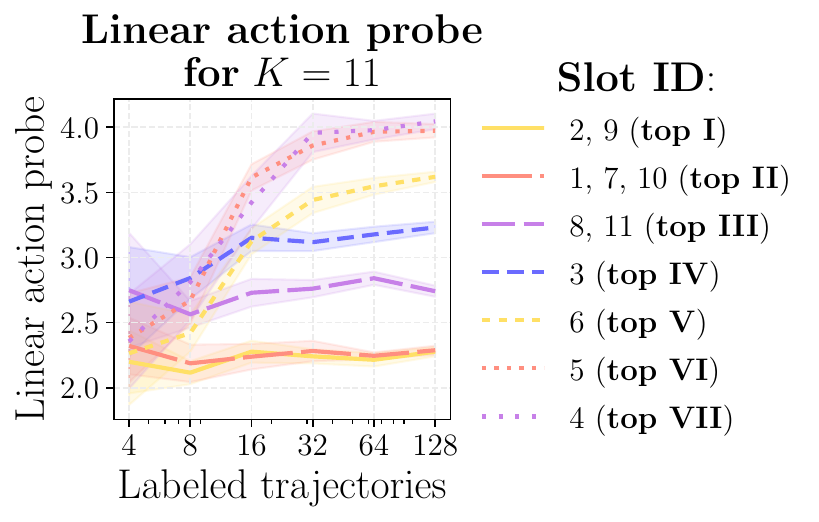}%
  \hfill
  \includegraphics[height=\myheightt]{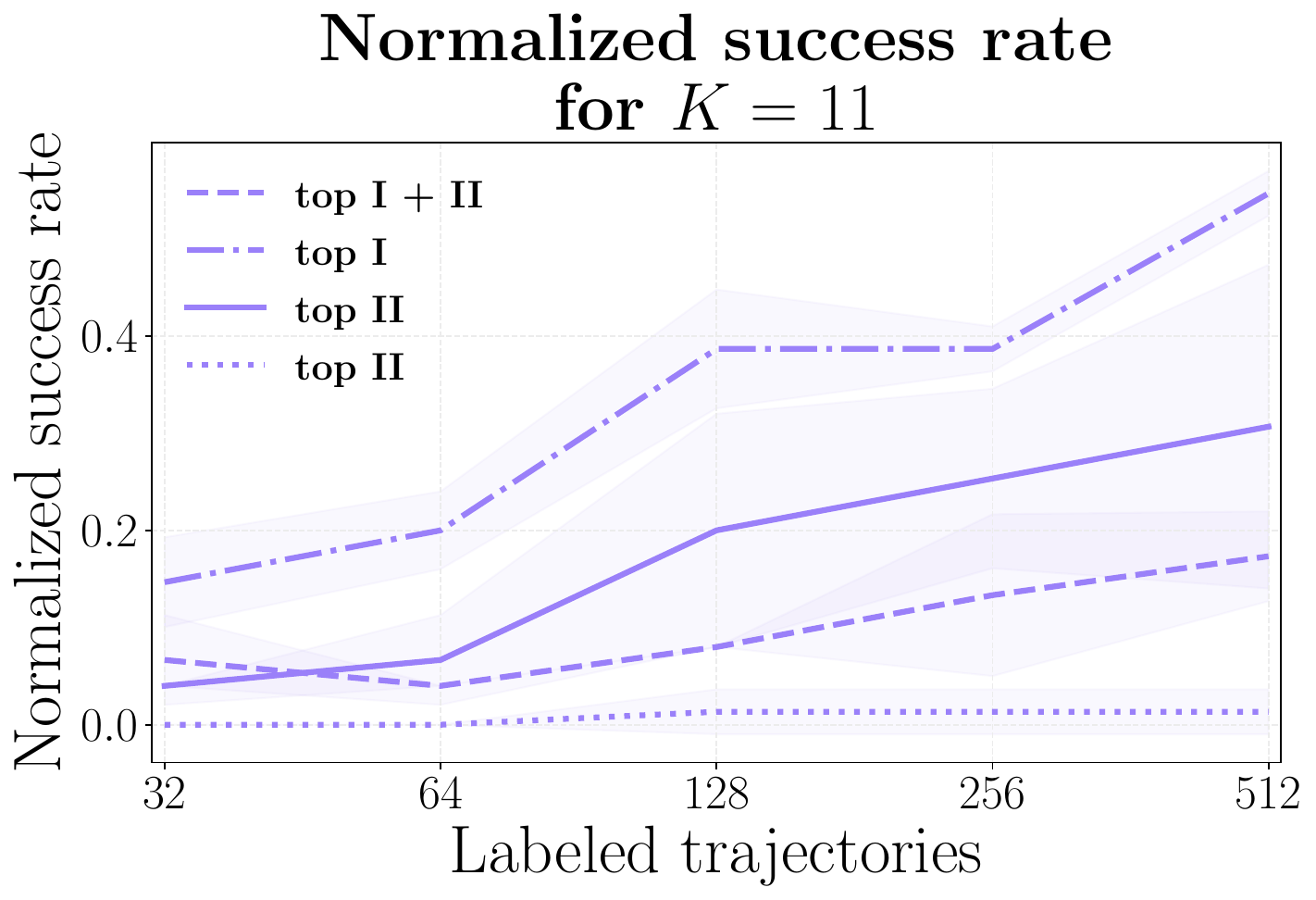}%
  \hfill
  \includegraphics[height=\myheight]{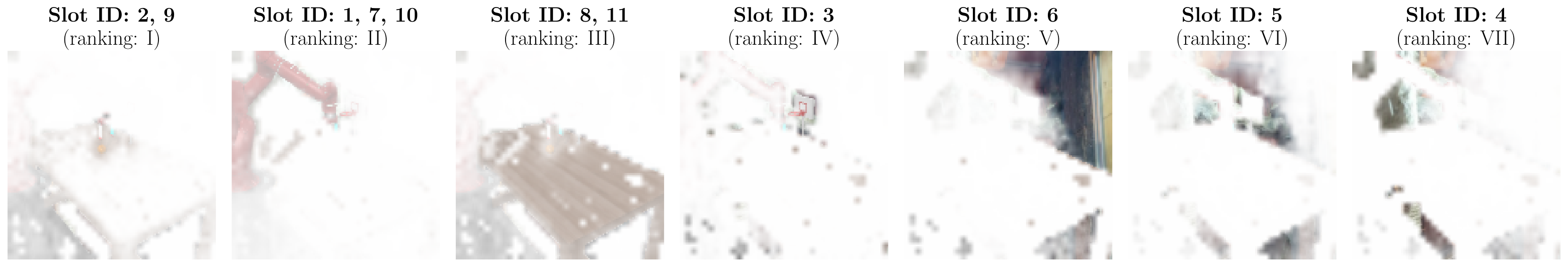}%
  
  \caption{Slot selection for \emph{11} slots. From left to right: (1) Linear action probes for VideoSaur slots pretrained basketball task for varying budget of trajectories. (2) Downstream performance of top I, top II, top III and concatenated top I+II slots pretrained basketball task for varying budget of trajectories. (3) Corresponding slot masks for basketball task after VideoSaur pretraining for number of slots $K=11$.}
  \label{fig:probe-corr-basketball-11}
\end{figure}

\begin{figure}[h!]
  \centering
  \def\myheightt{4cm}
  \def\myheight{2.6cm}
  
  \hfill
  \includegraphics[height=\myheightt]{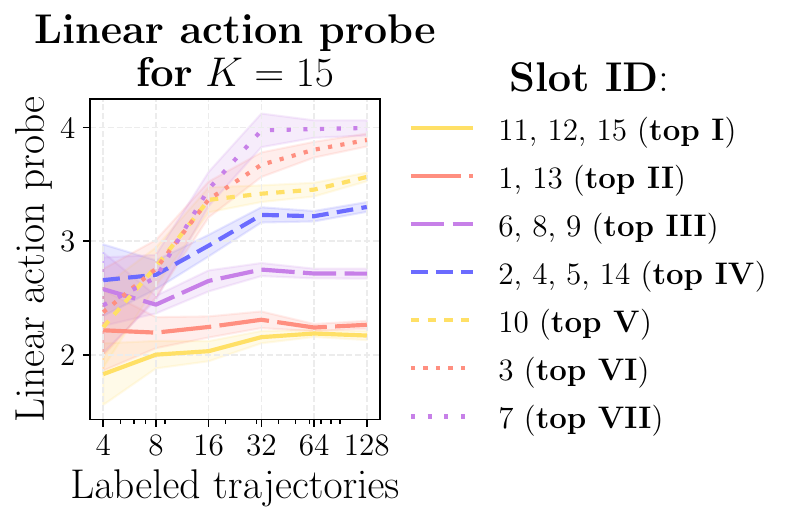}%
  \hfill
  \includegraphics[height=\myheightt]{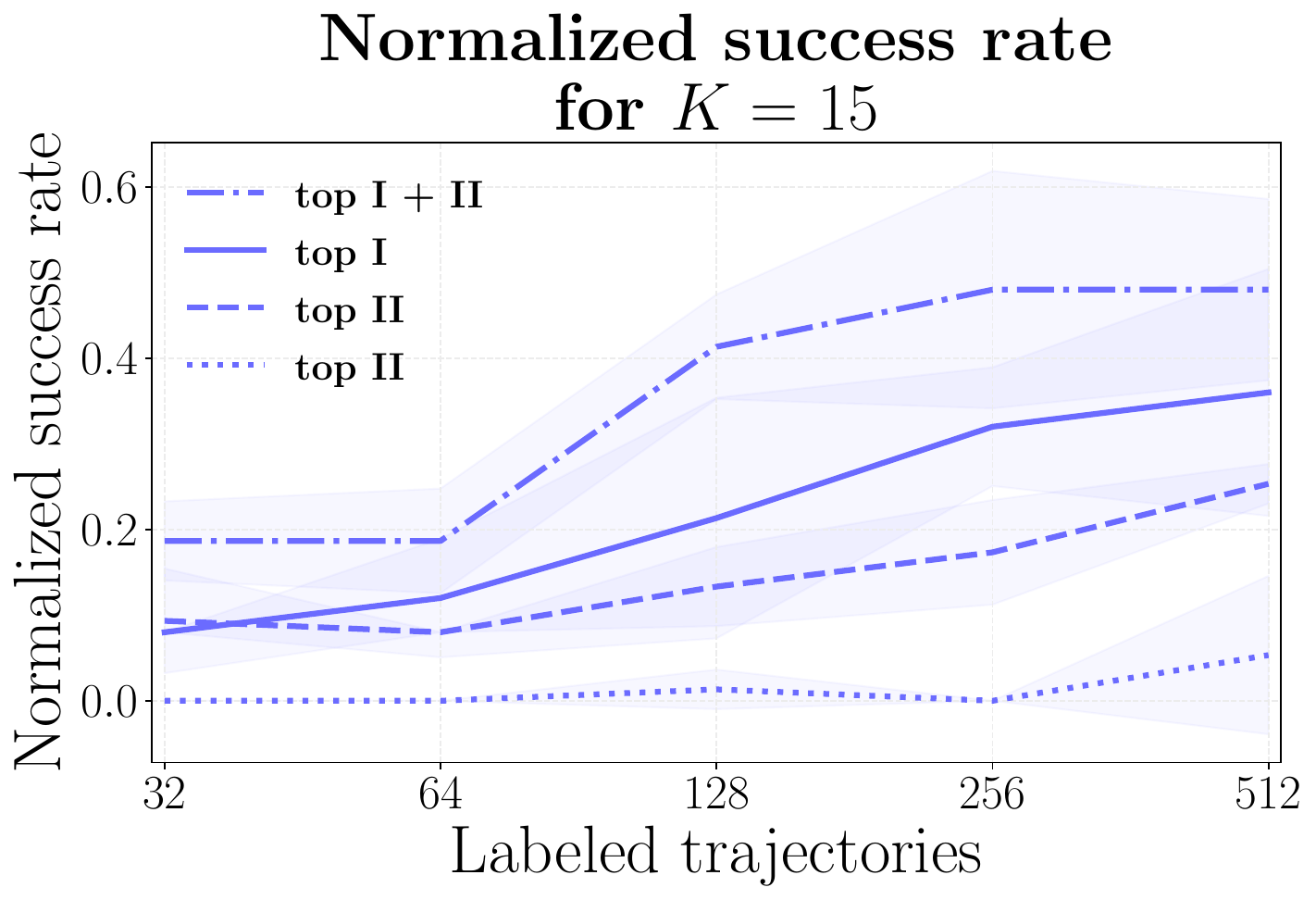}%
  \hfill \\
  \hfill
  \includegraphics[height=\myheight]{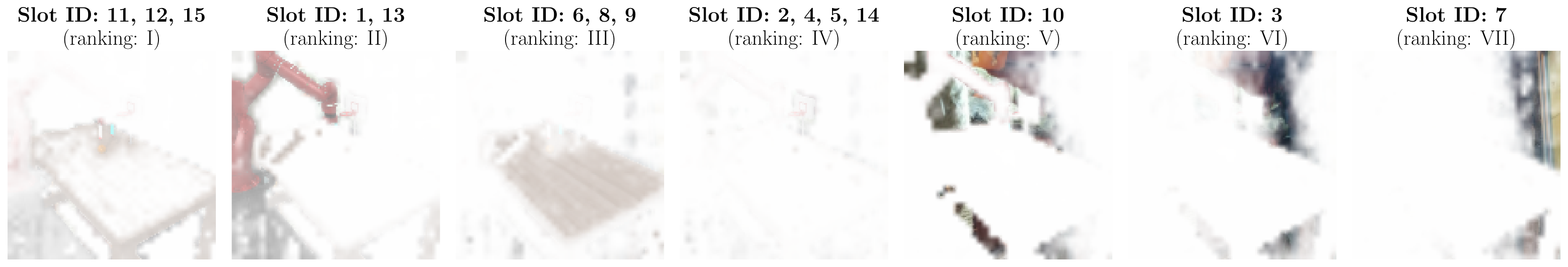}%
  
  \caption{Slot selection for \emph{15} slots. From left to right: (1) Linear action probes for VideoSaur slots pretrained basketball task for varying budget of trajectories. (2) Downstream performance of top I, top II, top III and concatenated top I+II slots pretrained basketball task for varying budget of trajectories. (3) Corresponding slot masks for basketball task after VideoSaur pretraining for number of slots $K=15$.}
  \label{fig:probe-corr-basketball-15}
\end{figure}

\newpage

\section{Ablation on the usage of DINO features}

\begin{figure}[h!]
\begin{center}
\includegraphics[width=0.4\textwidth]{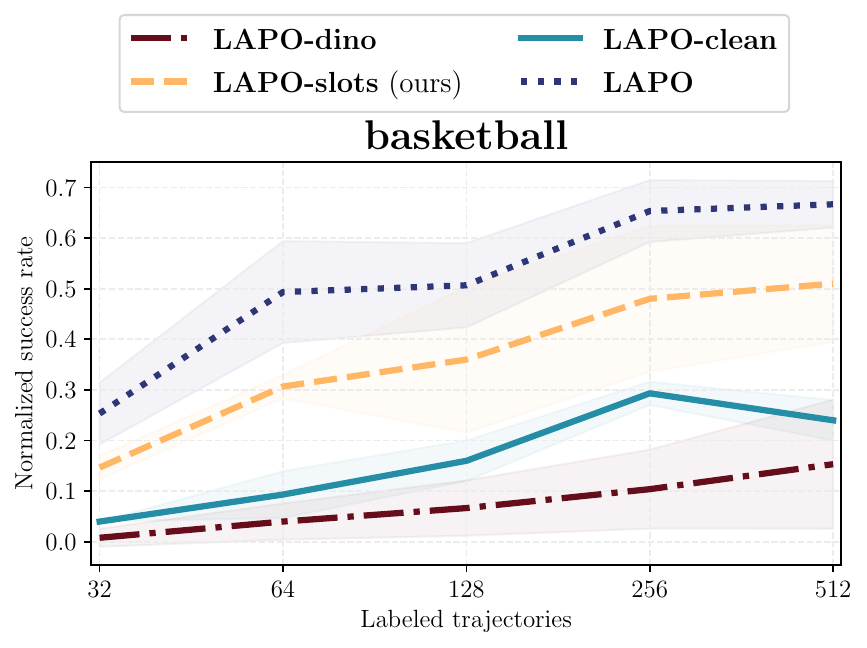}
\caption{Ablation study on the DMW basketball task to disentangle the effects of DINO features and object-centric slots. The plot compares four lines: LAPO (baseline), LAPO-clean (upper bound on clean data), LAPO-dino (LAPO based on DINO features), and LAPO-slots (our method, using DINO features within an object-centric framework).}
\label{fig:dino-ablation}
\end{center}
\end{figure}

Our primary method, LAPO-slots, leverages VideoSAUR, which utilizes a powerful DINOv2 pretrained encoder. This raises an important question: does the performance improvement stem from the object-centric slot representations, or simply from the stronger features provided by DINO? To disentangle these effects, we conduct an ablation study. We introduce a new baseline, \emph{LAPO-dino}, which uses features of DINOv2's CLS token from \cite{oquab2023dinov2}. This allows for a direct comparison between applying DINO features to raw pixels (LAPO-dino) versus applying them within our object-centric framework (LAPO-slots). The results, shown in \Cref{fig:dino-ablation}, demonstrate that DINO features alone do not yield a performance gain over standard LAPO, likely due to challenges specific to video processing.

\section{Comparing STEVE and VideoSAUR slot projections}
\label{app:steve}

\begin{figure}[h!]
\centering
\includegraphics[width=0.4\textwidth]{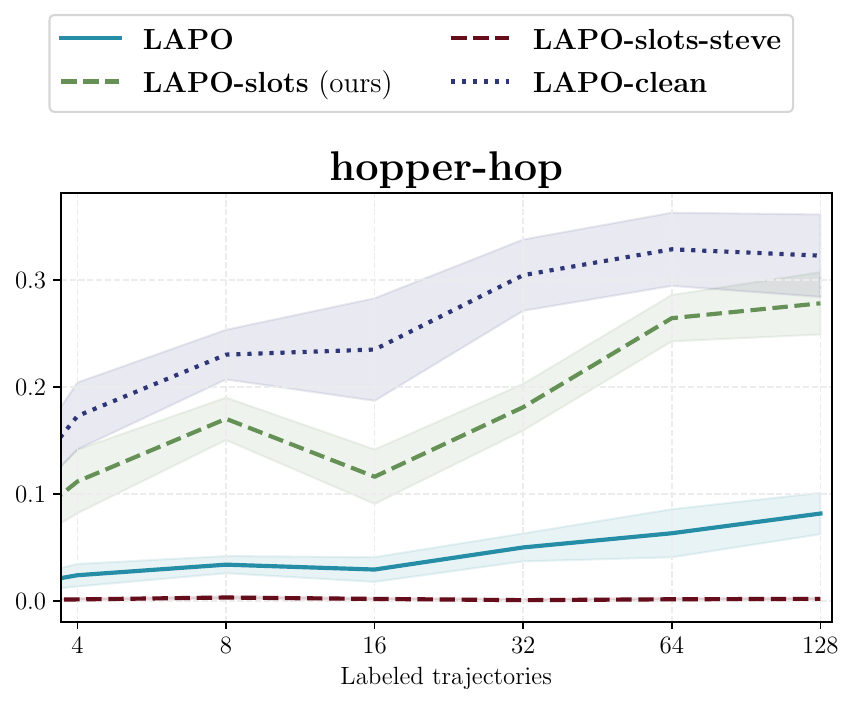}
\caption{Normalized evaluation returns the BC agent trained on latent actions for varying numbers of fine-tuning labeled trajectories. TL;DR: Object-centric pretraining based on STEVE doesn't work.}
\label{fig:steve-performance}
\end{figure}

\begin{figure}[h!]
\centering
\includegraphics[width=0.75\textwidth]{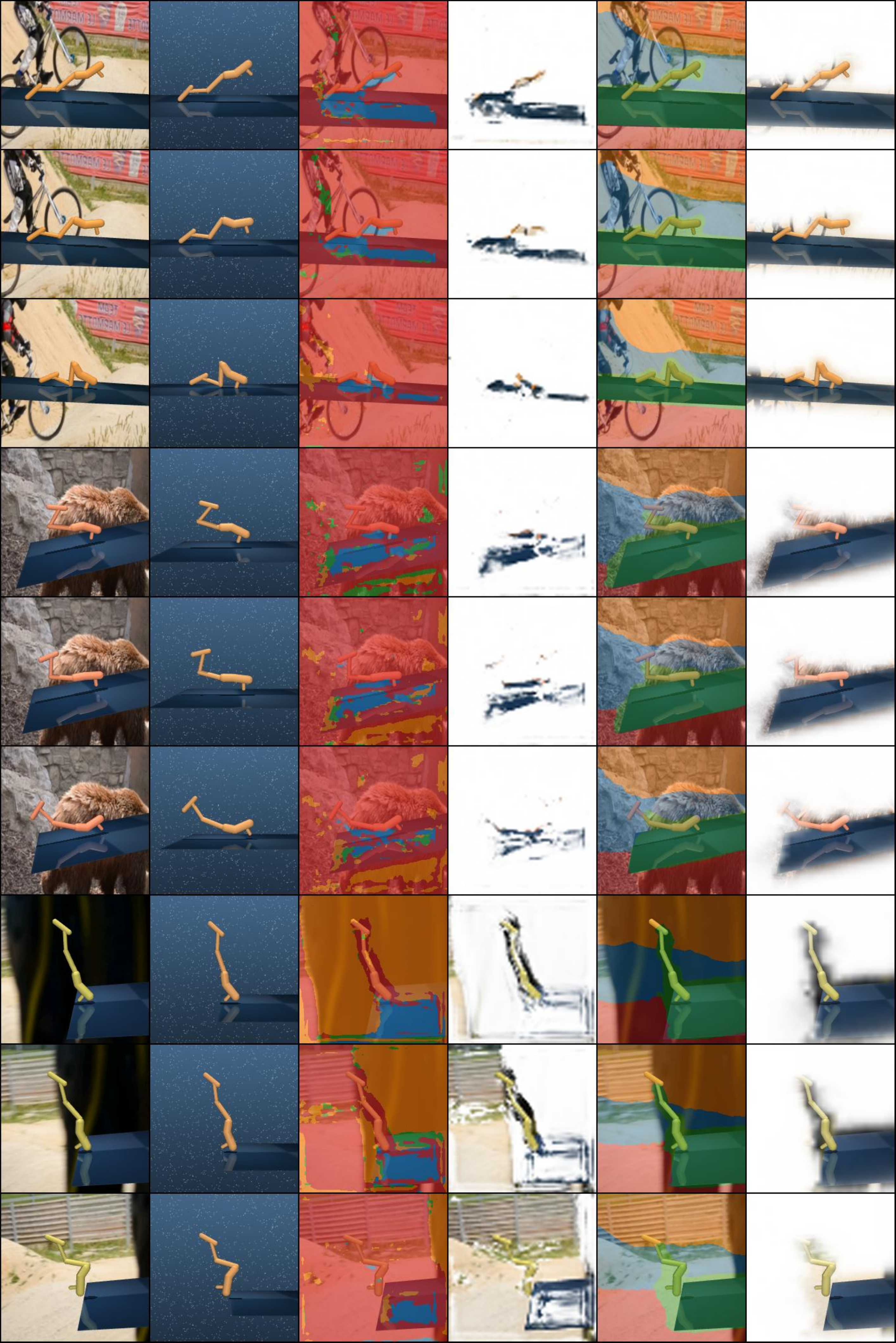}
\caption{Examples of slot projections on DCS hopper-hop. From left to right: clean image without distractors, image with distractors, STEVE's slots projections, the main object's slot projection by STEVE, VideoSAUR's slot projections, the main object's slot by VideoSAUR}
\label{plot:videosaur-VS-steve}
\end{figure}

We begin with the STEVE model \citep{singh2022simpleunsupervisedobjectcentriclearning}, a widely adopted and promising approach for object-centric learning. However, we found that STEVE struggled to accurately identify the main object in visual inputs. In \Cref{plot:videosaur-VS-steve}, we compare the slot projections of STEVE and VideoSAUR, highlighting key differences in their representations. We also note that STEVE is largely similar to SAVi \citep{kipf2022conditionalobjectcentriclearningvideo}, differing primarily in its use of a transformer-based decoder instead of a pixel-mixture decoder. In terms of performance, we evaluate the STEVE-based object-centric model on the hopper-hop task from DCS (\Cref{fig:steve-performance}), which underscores the importance of a strong object-centric representation for achieving good final performance.

\clearpage
\section{Examples of VideoSAUR slot projections on DCS-hard}
\begin{figure}[h!]
\begin{center}
\includegraphics[width=3.9in]{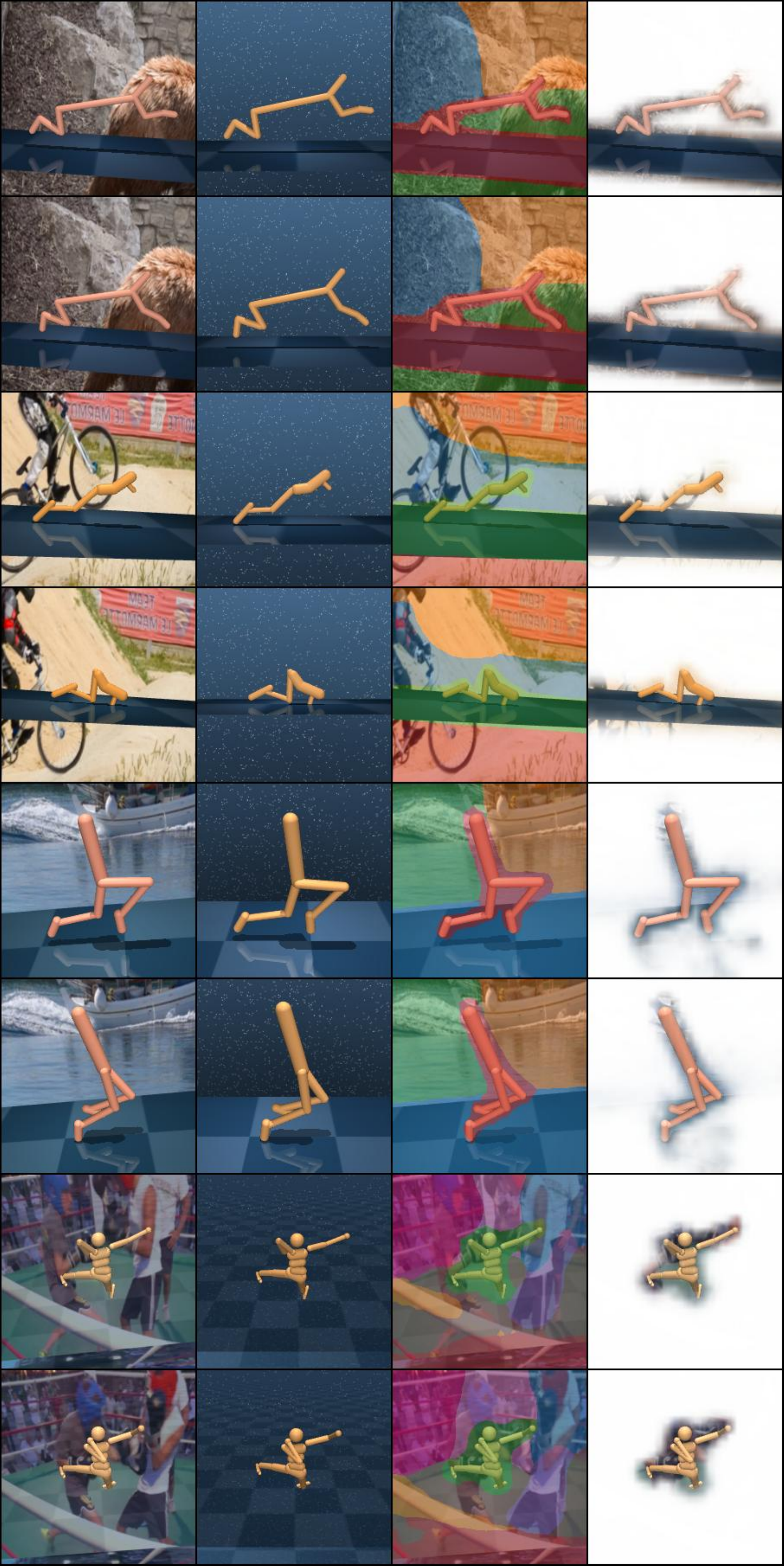}
\caption{Examples of VideoSAUR slot projections on DCS-hard for 4 tasks: (from upper to lower) cheetah-run, hopper-hop, walker-run, humanoid-walk. From left to right: distracted observation, clean observation, observation with segments of slot projections, slot projection (we call it "mask") of the main object}
\label{plot:video-saur-main-object-slots}
\end{center}
\end{figure}

\newpage
\begin{table}
\caption{Object-centric pretraining hyperparams. As for the number of slots: all DCS tasks were trained with 4 slots, except humanoid trained with 8 slots, all DMW tasks were trained with 3 slots for main tables.}
\label{table:hyper-ocl}
\begin{center}
\begin{tabular}{l|l}
\toprule
\bf Hyperparameter & \bf Value\\ 
\midrule
Episode Length & 3 \\ \hline
Image Size Dataset & 64 \\ \hline
Image Size Resize & 224 \\ \hline
Max Steps & 100000 \\ \hline
Batch Size & 256 \\ \hline
Warmup Steps & 2500 \\ \hline
Weight Decay & 0 \\ \hline
Max Video Length & 1000 \\ \hline
Gradient Clip Value & 0.05 \\ \hline
Slot Dimension & 128 \\ \hline
Vision Transformer Model & vit\_base\_patch8\_224\_dino \\ \hline
Feature Dimension & 768 \\ \hline
Number of Patches & 784 \\ \hline
Batch Size per GPU & 128 \\ \hline
Total Batch Size  & 128 \\ \hline
Similarity Temperature  & 0.075 \\ \hline
Similarity Weight  & 0.1 \\ \hline
Base Learning Rate  & 0.0001 \\ \hline
Learning Rate & 0.0003 \\ \hline
Learning Rate Scheduler & exp\_decay\_with\_warmup \\ \hline
Warmup Steps & 2500 \\ \hline
Decay Steps & 100000 \\ \hline
\bottomrule
\end{tabular}
\end{center}
\end{table}

\begin{table}
\caption{Hyperparameter optimization setups for latent action learning from vector representations (used for LAPO-slots).}
\label{table:hyper-lapo-slots}
\begin{center}
\begin{tabular}{l|l|l}
\toprule
\bf Hyperparameter & \bf DCS, DCS-Hard & \bf DMW\\  
\midrule
\multicolumn{3}{c}{\bf Latent Action Learning} \\ \hline
Batch Size & 8192 & 64 \\ \hline
Hidden Dimension & 1024 & 1024 \\ \hline
Num Blocks & 3 & [3, 5, 8] \\ \hline
Number of Epochs & 30 & 30 \\ \hline
Frame Stack & [1, 3] & [1, 3] \\ \hline
Weight Decay & 0 & 0 \\ \hline
Learning Rate & log(1e-3, 1e-06) & log(1e-3, 1e-06) \\ \hline
Warmup Epochs & 3 & 3 \\ \hline
Future Observation Offset & 10 & 10 \\ \hline
Latent Action Dimension & 8192 & 512 \\ \hline
\midrule
\multicolumn{3}{c}{\bf BC} \\ \hline
Dropout & 0 & 0 \\ \hline
Use Augmentation & False & False \\ \hline
Evaluation Seed  & 0 & 0 \\ \hline
Batch Size & 512 & 64 \\ \hline
Number of Epochs & 10 & 10 \\ \hline
Frame Stack & 3 & 3 \\ \hline
Encoder Deep & False & True \\ \hline
Weight Decay & 0 & 0 \\ \hline
Encoder Scale & 32 & 8 \\ \hline
Evaluation Episodes & 5 & 5 \\ \hline
Learning Rate & 0.0001 & 0.0001 \\ \hline
Warmup Epochs & 0 & 0 \\ \hline
Encoder Number of Residual Blocks & 2 & 2 \\ \hline
\midrule
\multicolumn{3}{c}{\bf BC finetuning} \\ \hline
Use Augmentation & False & False \\ \hline
Batch Size & 512 & 64 \\ \hline
Hidden Dimension & 256 & [64, 512] \\ \hline
Weight Decay & 0 & 0 \\ \hline
Evaluation Episodes & 25 & 25 \\ \hline
Learning Rate & 0.0003 & 0.0001 \\ \hline
Total Updates & 2500 & 15000 \\ \hline
Warmup Epochs & 0 & 5 \\ \hline
\bottomrule
\end{tabular}
\end{center}
\end{table}

\begin{table}
\caption{Hyperparameter optimization setup for latent action learning from images (used for LAPO, LAPO-masks, LAPO-clear).}
\label{table:hyper-lapo-masks}
\begin{center}
\begin{tabular}{l|l|l}
\toprule
\bf Hyperparameter & \bf DCS & \bf DMW\\  
\midrule
\multicolumn{3}{c}{\bf Latent Action Learning} \\ \hline
Batch Size & 512 & 64 \\ \hline
Number of Epochs & 10 & 10 \\ \hline
Frame Stack & 3 & 3 \\ \hline
Encoder Deep & False & True \\ \hline
Weight Decay & 0 & 0 \\ \hline
Encoder Scale & 6 & 4 \\ \hline
Learning Rate & log(1e-3, 1e-06) & log(1e-3, 1e-06) \\ \hline
Warmup Epochs & 3 & 3 \\ \hline
Future Observation Offset & 10 & 10 \\ \hline
Latent Action Dimension & 1024 & 1024 \\ \hline
Encoder Number of Residual Blocks & 2 & [1, 2] \\ \hline
\midrule
\multicolumn{3}{c}{\bf BC} \\ \hline
Dropout & 0 & 0 \\ \hline
Use Augmentation & False & False \\ \hline
Evaluation Seed & 0 & 0 \\ \hline
Batch Size & 512 & 64 \\ \hline
Number of Epochs & 10 & 10 \\ \hline
Frame Stack & 3 & 3 \\ \hline
Encoder Deep & False & False \\ \hline
Weight Decay & 0 & 0 \\ \hline
Encoder Scale & 32 & 8 \\ \hline
Evaluation Episodes & 5 & 5 \\ \hline
Learning Rate & 0.0001 & 0.0001 \\ \hline
Warmup Epochs & 0 & 0 \\ \hline
Encoder Number of Residual Blocks & 2 & [1,2] \\ \hline
\midrule
\multicolumn{3}{c}{\bf BC finetuning} \\ \hline
Use Augmentation & False & False \\ \hline
Evaluation Seed & 0 & 0 \\ \hline
Batch Size & 512 & 64 \\ \hline
Hidden Dimension & 256 & [64, 512] \\ \hline
Weight Decay & 0 & 0 \\ \hline
Evaluation Episodes & 25 & 25 \\ \hline
Learning Rate & 0.0003 & 0.0001 \\ \hline
Total Updates & 2500 & 15000 \\ \hline
Warmup Epochs & 0 & 0 \\ \hline
\bottomrule
\end{tabular}
\end{center}
\end{table}




\end{document}